\documentclass[runningheads]{llncs}

\usepackage{eccv}

\usepackage{eccvabbrv}

\usepackage{graphicx}
\usepackage{booktabs} 
\usepackage{pifont}

\usepackage[accsupp]{axessibility}  
\usepackage[percent]{overpic}
\usepackage{amsmath}

\usepackage{bm}
\usepackage{colortbl}
\usepackage{multirow}
\usepackage{collcell}
\usepackage{subcaption}
\usepackage{array}
\usepackage{arydshln}
\usepackage{algorithm}
\usepackage{algpseudocode}
\usepackage{algorithmicx}
\usepackage{listings}
\usepackage{enumitem}
\usepackage{etoolbox}
\usepackage{caption}

\usepackage{wrapfig}
\usepackage{tikz} 
\usepackage{xcolor}
\definecolor{codegreen}{RGB}{79,126,127}
\definecolor{codedefine}{RGB}{153,54,159}
\definecolor{codefunc}{RGB}{73,122,234}
\definecolor{codecall}{RGB}{73,122,234}
\definecolor{codepro}{RGB}{212,96,80}
\definecolor{codedim}{RGB}{89,152,195}
\definecolor{codeyellow}{RGB}{255, 215, 0}
\definecolor{mycodeblue}{RGB}{70, 130, 180}

\makeatletter
\AfterEndEnvironment{algorithm}{\let\@algcomment\relax}
\AtEndEnvironment{algorithm}{\kern2pt\hrule\relax\vskip3pt\@algcomment}
\let\@algcomment\relax
\newcommand\algcomment[1]{\def\@algcomment{\footnotesize#1}}
\renewcommand\fs@ruled{\def\@fs@cfont{\bfseries}\let\@fs@capt\floatc@ruled
  \def\@fs@pre{\hrule height.8pt depth0pt \kern2pt}%
  \def\@fs@post{}%
  \def\@fs@mid{\kern2pt\hrule\kern2pt}%
  \let\@fs@iftopcapt\iftrue}
\makeatother

\makeatletter
\newcommand{\thickhline}{%
    \noalign {\ifnum 0=`}\fi \hrule height 1pt
    \futurelet \reserved@a \@xhline
}

\newcommand{\spthickhline}{%
    \noalign {\ifnum 0=`}\fi \hrule height 1.5pt
    \futurelet \reserved@a \@xhline
}
\newcommand{\sspthickhline}{%
    \noalign {\ifnum 0=`}\fi \hrule height 1.6pt
    \futurelet \reserved@a \@xhline
}
\definecolor{lightred}{RGB}{255,204,203}

\newcommand{\maketitlesupplementary}{%
  \clearpage
  {%
    \centering
    {\Large\bfseries PKINet-v2: Towards Powerful and Efficient Poly-Kernel Remote Sensing Object Detection\par}
    \vspace{0.5em}
    {\large Supplementary Material\par}
    \vspace{1.0em}
  }%
  \normalsize
  \raggedright
}

\usepackage[pagebackref,breaklinks,colorlinks,citecolor=eccvblue]{hyperref}

\usepackage{orcidlink}

\begin{document}

\title{PKINet-v2: Towards Powerful and Efficient Poly-Kernel Remote Sensing Object Detection} 

\titlerunning{PKINet-v2}

\author{
\makebox[\textwidth][c]{%
\begin{tabular}{>{\centering\arraybackslash}m{0.31\textwidth}
                >{\centering\arraybackslash}m{0.31\textwidth}
                >{\centering\arraybackslash}m{0.31\textwidth}}
Xinhao Cai$^{1,4}$ & Liulei Li$^{2}$ & Gensheng Pei$^{3}$ \\
Zeren Sun$^{1,4}$   & Yazhou Yao$^{1,4,\dagger}$ & Wenguan Wang$^{2,\dagger}$
\end{tabular}%
}
}

\authorrunning{X.~Cai et al.}

\institute{
\parbox{\textwidth}{\centering
${}^{1}$ Nanjing University of Science and Technology \qquad
${}^{2}$ Zhejiang University\\
${}^{3}$ Department of Electrical and Computer Engineering, Sungkyunkwan University\\
${}^{4}$ State Key Lab. of Intelligent Manufacturing of Advanced Construction Machinery
}
}

\authorrunning{X.~Cai et al.}

\maketitle

\begingroup
\renewcommand{\thefootnote}{}
\footnotetext{${}^{\dagger}$ Corresponding authors.}
\endgroup

\vspace{-10pt}
\begin{center}
\url{https://github.com/NUST-Machine-Intelligence-Laboratory/PKINet}
\end{center}

\begin{abstract}
    Object detection in remote sensing images (RSIs) is challenged by the coexistence of geometric and spatial complexity: targets may appear with diverse aspect ratios, while spanning a wide range of object sizes under varied contexts. Existing RSI backbones address the two challenges separately, either by adopting anisotropic strip kernels to model slender targets or by using isotropic large kernels to capture broader context. However, such isolated treatments lead to complementary drawbacks: the strip-only design can disrupt spatial coherence for regular-shaped objects and weaken tiny details, whereas isotropic large kernels often introduce severe background noise and geometric mismatch for slender structures. In this paper, we extend PKINet, and present a powerful and efficient backbone that jointly handles both challenges within a unified paradigm named Poly Kernel Inception Network v2 (PKINet-v2). PKINet-v2 synergizes anisotropic axial-strip convolutions with isotropic square kernels and builds a multi-scope receptive field, preserving fine-grained local textures while progressively aggregating long-range context across scales. To enable efficient deployment, we further introduce a Heterogeneous Kernel Re-parameterization (HKR) Strategy that fuses all heterogeneous branches into a single depth-wise convolution for inference, eliminating fragmented kernel launches without accuracy loss. Extensive experiments on four widely-used benchmarks, including DOTA-v1.0, DOTA-v1.5, HRSC2016, and DIOR-R, demonstrate that PKINet-v2 achieves state-of-the-art accuracy while delivering a $\textbf{3.9}\times$ FPS acceleration compared to PKINet-v1, surpassing previous remote sensing backbones in both effectiveness and efficiency.
\end{abstract}

\section{Introduction}
\label{sec:intro}
Remote sensing images~\cite{cheng2014multi} (RSIs), comprising aerial and satellite photography, are captured from a bird's-eye perspective, offering high-resolution views of the Earth's surface. Object detection in RSIs~\cite{xia2018dota, ding2021object, sun2022fair1m}, which aims to identify specific targets and determine spatial coordinates and class labels, has garnered significant attention in recent years due to its wide range of applications, including urban planning, environmental monitoring, and disaster management~\cite{zhang2016deep, ma2019deep}. 

In contrast to generic object detection in natural scenes, object detection in RSIs presents two primary challenges: \ding{182} \textbf{Geometric Complexity}: Targets often appear with \textit{arbitrary orientations} and \textit{diverse aspect ratios} (\textit{e.g.}, from roundabouts to bridges); \ding{183} \textbf{Spatial Complexity}: Objects exhibit \textit{large-scale variations}, ranging from soccer fields to tiny vehicles, coupled with \textit{wide-ranging contexts} that demand capturing both fine-grained details and global context.

\begin{figure*}[t!]
\vspace{-10pt}
    \centering    \includegraphics[width=0.85\linewidth]{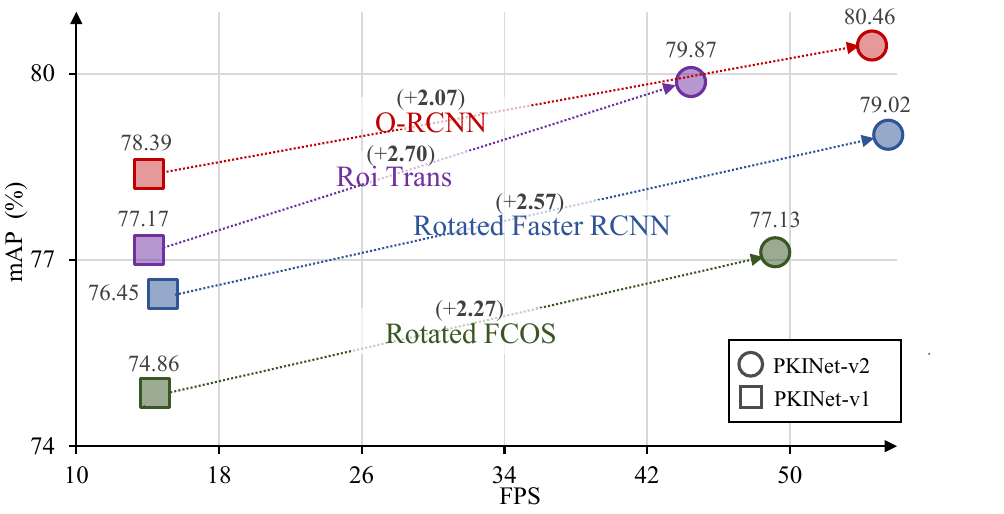}
        \put(-134,50.1){\scriptsize\cite{tian2019fcos}}
        \put(-99,82){\scriptsize\cite{ren2015faster}}
        \put(-154,113){\scriptsize\cite{xie2021oriented}}
        \put(-165,96.5){\scriptsize\cite{ding2019learning}}
        \put(-42,32){\scriptsize\cite{cai2024poly}}
    \vspace{-10pt}
    \caption{\textbf{Performance of PKINet-v2 on DOTA-v1 dataset}~\cite{xia2018dota}. PKINet-v2 consistently improves mAP while substantially boosting inference efficiency over PKINet-v1~\cite{cai2024poly}, across representative oriented detectors.
    }
    \vspace{-20pt}
    \label{fig:performance}
\end{figure*}
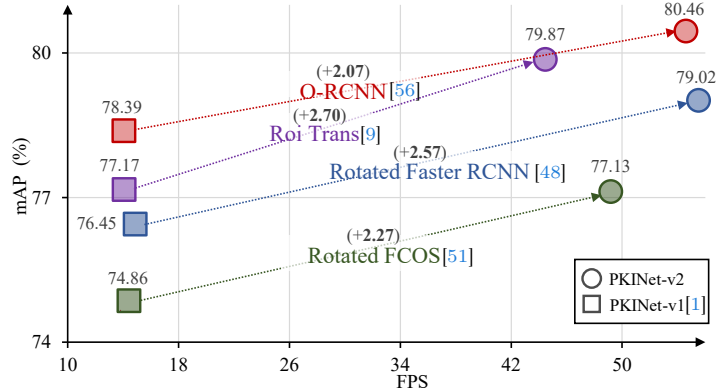

To mitigate \ding{182}, extensive prior efforts have been dedicated to developing robust Oriented Bounding Box (OBB) detectors~\cite{xu2020gliding, li2022oriented, xie2021oriented, ding2019learning, yang2021r3det, han2021align} and refining angle prediction accuracy for OBB~\cite{yang2021dense, yang2020arbitrary,yang2021rethinking, yang2021learning, yang2022kfiou}. From the feature extraction perspective, recently, Strip RCNN~\cite{yuan2025strip} introduces strip kernels to replace standard square convolutions, thereby modeling the geometry of objects with extreme aspect ratios. Nevertheless, Strip RCNN's exclusive reliance on strip kernels disrupts the spatial coherence of isotropic objects (\textit{e.g.}, storage tanks and roundabouts) and dilutes the fine-grained details of tiny instances. To mitigate \ding{183}, LSKNet~\cite{Li_2023_ICCV} employs a spatial selection mechanism with large kernels to dynamically adjust the receptive field for varying contexts. However, LSKNet is constrained by its isotropic square large kernels.
While effective for capturing global context, the geometric mismatch with slender objects leads 
to severe background noise interference. More critically, these methods overlook the extreme variations in object scales inherent in RSIs.

To address substantial scale variations in RSIs, PKINet-v1~\cite{cai2024poly} adopts parallel convolutions with diverse kernel sizes to construct a multi-scale receptive field, while concurrently deploying a Context Anchor Attention mechanism to encompass long-range contextual information. Despite these advances, PKINet-v1 relies solely on isotropic square convolutions, which limit its geometric adaptability. Moreover, its intricate multi-branch architecture imposes memory fragmentation during inference, resulting in suboptimal real-time inference speed.

\begin{wraptable}{r}{0.48\textwidth}
  \vspace{-10pt}
    \centering
    \caption{\textbf{Remote Sensing Backbones Comparison}. All backbones are pretrained on ImageNet-1K \cite{deng2009imagenet} for 300 epochs and built within the framework of Oriented RCNN~\cite{xie2021oriented}. FPS is tested on a single NVIDIA A100-40G GPU.}
    \setlength{\tabcolsep}{3pt}
    \renewcommand\arraystretch{1.15}
    \label{tab:fps_comparison}
    \resizebox{\linewidth}{!}{
        \begin{tabular}{r|c||c|c|c}
            \hline
            \rowcolor[rgb]{0.92,0.92,0.92}\textbf{Backbone}~~~~~ & \textbf{mAP} $\uparrow$ & \textbf{FPS} $\uparrow$ & \#\textbf{P} $\downarrow$ & \textbf{FLOPs} $\downarrow$ \\
            \hline
            \hline
            LSKNet-S~\cite{Li_2023_ICCV}     & 77.49 & 51.00 & 31.0M & 161G \\
            StripNet-S~\cite{yuan2025strip} & 79.85 & 46.40 & 30.5M & 172G \\
            LWGANet-L2~\cite{lu2025lwganet}   & 79.02 & 33.50 & 29.2M & 159G \\
            PKINet-v1-S~\cite{cai2024poly}     & 78.39 & 14.05 & 30.8M & 184G \\
            \rowcolor[rgb]{0.92,0.92,0.92} \textbf{PKINet-v2-S}~~~    & \textbf{80.46} & 54.60 & 30.7M & 173G \\
            \hline
        \end{tabular}
    }
    \vspace{-20pt}
\end{wraptable}  
To simultaneously address \ding{182} geometric complexity and \ding{183} spatial complexity challenges within RSIs, in this paper, we present a powerful and efficient feature extraction backbone network named Poly Kernel Inception Network v2 (PKINet-v2) for remote sensing object detection.
\textbf{\textit{First}}, to tackle \ding{182},  distinct 
 from previous methods that restrict themselves to homogeneous kernel designs by utilizing either solely strip convolutions or square kernels, PKINet-v2 synergizes anisotropic strip convolutions with isotropic square convolutions. This hybrid approach ensures robust modeling for both slender targets and regular-shaped objects. \textbf{\textit{Second}}, to tackle \ding{183}, PKINet-v2 integrates hierarchically densified multi-scale receptive fields spanning from dense small kernels to expansive context windows. This design enables the network to meticulously preserve local fine-grained details while simultaneously aggregating long-range contextual information, thereby accommodating objects of extreme scale variations. \textbf{\textit{Third}}, to achieve high efficiency despite the complex model design, PKINet-v2 introduces a Heterogeneous Kernel Re-parameterization (HKR) Strategy. Specifically, it transforms the multi-branch training architecture into a hardware-friendly single-branch structure during inference. HKR removes memory fragmentation and synchronization costs, guaranteeing high-speed inference without accuracy degradation.

To the best of our knowledge, PKINet-v2 is the first attempt to explore a unified paradigm that simultaneously addresses the dual challenges of geometric and spatial complexity. The contributions are as follows:
\vspace{-6pt}
\begin{itemize}
    \item  We propose the Poly Kernel Inception Network v2 (PKINet-v2), a backbone that synergizes anisotropic strip convolutions with isotropic square convolutions. This unified kernel design targets the geometric complexity, enabling robust modeling of targets with diverse geometries, ranging from slender instances to regular-shaped objects.
    
    \item We construct a hierarchically densified multi-scale receptive field, integrating diverse kernels spanning from dense small filters to expansive context windows. This design targets the spatial complexity  by concurrently preserving local fine-grained details and aggregating long-range contextual information.
    
    \item We introduce a Heterogeneous Kernel Re-parameterization (HKR) Strategy to reconcile model complexity with inference efficiency. It transforms the multi-branch training architecture into a hardware-friendly single-branch structure during inference, eliminating memory fragmentation and guaranteeing high-speed inference without any accuracy degradation.
\end{itemize}

PKINet-v2 achieves state-of-the-art performance on four widely-used remote sensing benchmarks, namely DOTA-v1.0 \cite{xia2018dota}, DOTA-v1.5 \cite{xia2018dota}, HRSC2016 \cite{liu2017high}, and DIOR-R \cite{cheng2022anchor}, validating its effectiveness. Furthermore, in terms of efficiency, PKINet-v2 achieves a $\textbf{3.9}\times$ FPS acceleration compared to PKINet-v1 and surpasses previous remote sensing backbones, as reported in Tab.~\ref{tab:fps_comparison}.

\section{Related Work}
\label{sec:related}
As illustrated in \S\ref{sec:intro}, the challenges faced by remote sensing object detection
primarily stem from geometric and spatial complexity. To tackle the geometric challenge, extensive prior efforts were dedicated to robust Oriented Bounding Box (OBB) detectors. Nonetheless, a recent paradigm shift focuses on constructing effective feature extraction backbones explicitly tailored to the characteristics of remote sensing images (RSIs) to address the two challenges respectively.

\noindent\textbf{OBB for Remote Sensing Object Detection.}
To address the arbitrary orientations inherent in RSIs, existing literature can be broadly categorized into two streams: \textit{architectural adaptation} and \textit{representation reformulation}. 
The former integrates orientation-aware modules into detectors, such as introducing feature refinement techniques into the detector neck~\cite{yang2019scrdet,yang2021r3det}, refining features via rotated region of interest (RoI) extraction~\cite{ding2019learning,xie2021oriented}, and engineering alignment-friendly detection heads~\cite{yang2019scrdet,han2021redet,ming2021dynamic}. 
The latter seeks to fundamentally alleviate the boundary discontinuity issues stemming from angular periodicity by proposing novel representation schemes~\cite{xu2020gliding, li2022oriented, wang2019mask, yi2021oriented, fu2020point, yang2021dense}. 
For instance, Xu~\textit{et al}.~\cite{xu2020gliding} formulated a gliding vertex representation, which characterizes multi-oriented objects by introducing four gliding offset variables to the standard horizontal box. 
Adopting a discrete perspective, Li~\textit{et al}.~\cite{li2022oriented} represented oriented instances using a set of keypoints to enable more precise orientation estimation. 
Furthermore, statistical approaches have gained traction, where several works~\cite{hou2023g, yang2021rethinking, yang2021learning, cheng2022dual} model OBBs as Gaussian distributions, often incorporating tailored loss functions~\cite{qian2021learning} to guide the training process.

Despite their effectiveness in refining OBB predictions, these methods largely \textbf{focus on the detection head while overlooking the backbone}: they rely on generic backbones designed for natural images. Consequently, they neglect the unique \textit{geometric complexity} and \textit{spatial complexity} of RSIs during the fundamental feature extraction stage.

\noindent\textbf{Feature Extraction for Remote Sensing Object Detection.}
To address the spatial complexity challenge inherent in RSIs, most existing feature extraction methods have focused on handling large-scale variations.
These approaches typically employ strategies such as data augmentation~\cite{zhao2019multi, shamsolmoali2021rotation, chen2020stitcher}, multi-scale feature integration~\cite{zhang2019hierarchical,zheng2020hynet,9382268,liu2016ssd}, Feature Pyramid Network (FPN) enhancements~\cite{lin2017feature,hou2022refined,zhang2021laplacian, guo2020rotational}, and multi-scale anchor generation~\cite{guo2018geospatial, hou2021self, qiu2019a2rmnet}. More recently, growing attention has been directed towards the importance of \textit{contextual information} in RSIs. Meanwhile, there has been a shift towards \textit{designing specialized backbones} for remote sensing object detection. For instance, LSKNet~\cite{Li_2023_ICCV} employs a spatial selection mechanism with large kernels to dynamically adjust the receptive field for varying contexts. Subsequently, PKINet-v1~\cite{cai2024poly} adopts parallel convolutions with diverse kernel sizes to construct a multi-scale receptive field, while concurrently deploying a Context Anchor Attention mechanism to encompass long-range contextual information. LWGANet~\cite{lu2025lwganet} addresses spatial and channel redundancies via a lightweight grouped attention mechanism that decouples multi-scale features.

To address the geometric complexity challenge inherent in RSIs, previous methods focus on feature alignment strategies to resolve the mismatch between standard features and oriented objects. A recent paradigm shift directs attention towards designing geometry-aware backbones capable of extracting robust geometric features directly from the input. Several works~\cite{han2021redet,pu2023adaptive} aim to extract rotation-robust features by utilizing equivalent receptive fields for objects with varying orientations. In addition to orientation, recently, Strip RCNN~\cite{yuan2025strip} introduces strip convolutions to address the challenge of detecting slender objects.

In line with robust backbone design for RSIs, we propose a new feature extraction backbone, PKINet-v2, which tackles both geometric and spatial complexity simultaneously, distinct from prior methods that typically address these challenges in isolation. As an evolution of PKINet-v1~\cite{cai2024poly}, PKINet-v2 inherits the parallel multi-scale receptive field design while breaking the constraint of utilizing solely isotropic square kernels. By integrating anisotropic strip convolutions with isotropic square kernels, it establishes a hybrid receptive field, enabling robust modeling of both slender and compact objects. Furthermore, PKINet-v2 incorporates a Heterogeneous Kernel Re-parameterization (HKR) Strategy to ensure high-efficiency inference without compromising accuracy.
Remarkably, it achieves a performance gain of \textbf{2.07}\% mAP while accelerating the inference speed by $\textbf{3.9}\times$ compared to the PKINet-v1.

\section{Methodology} \label{sec:method}

\subsection{Limitations in PKINet-v1}
PKINet-v1~\cite{cai2024poly} successfully pioneered the use of multi-scale kernels to address scale variations in RSIs. However, despite its success, we identify three critical limitations that hinder its efficiency and effectiveness in general scenarios:

\begin{itemize}
    \item \textbf{Geometric Mismatch with Slender Objects:} PKINet-v1 utilizes solely \textit{dense isotropic square kernels} (\textit{i.e.}, $3 \times 3$ to $11 \times 11$), which are ill-suited for \textit{slender objects} common in remote sensing (\textit{e.g.}, bridges). This mismatch introduces excessive background noise and fails to capture the intrinsic geometry of elongated targets, compromising feature representation.
    
    \item \textbf{Computational Redundancy:} The reliance on large dense kernels incurs \textit{quadratic complexity}. As receptive fields expand, the quadratic increase in computational overhead results in slow convergence and limited throughput.
    
    \item \textbf{Fragmented Inference and Latency:} During inference, the independent operation of multi-branch kernels leads to \textit{fragmented execution}, characterized by frequent kernel launches and repeated memory access. This inefficiency severely hampers GPU parallel utilization, resulting in high latency.
\end{itemize}

\subsection{Overall Architecture of PKINet-v2}
\begin{figure*}[t!]
\vspace{-10pt}
    \centering    \includegraphics[width=1\linewidth]{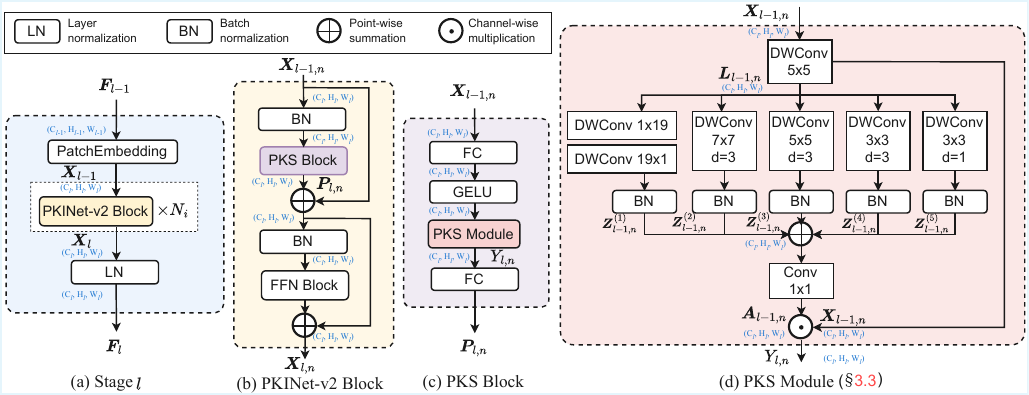}
    \vspace{-18pt}
    \caption{\textbf{PKINet-v2 overview.} PKINet-v2 consists of four stages. Each \textbf{(a)} \textbf{Stage $l$} contains a Patch Embedding layer followed by a sequence of $N_l$ PKINet-v2 Blocks. Each \textbf{(b) PKINet-v2 Block} is composed of a PKS sub-block and an FFN sub-block. The \textbf{(c) PKS Block} mainly comprises a PKS Module and two Fully Connected (FC) layers. \textbf{(d) PKS Module} (\S\ref{sec:PKS}) builds a multi-scope spatial attention map by aggregating heterogeneous branches with wide-range receptive field and fusing them with a $1\times1$ convolution. Here, $n\!=\!0,\dots,N_l\!-\!1$ indicates that the PKS Module/Block is located in the $n$-th PKINet-v2 Block of the $l$-th stage. Please refer to \S\ref{sec:method} for more details.
    }
    \vspace{-16pt}
    \label{fig:framework}
\end{figure*}

As illustrated in Fig.~\ref{fig:framework}, similar to PKINet-v1~\cite{cai2024poly}, our PKINet-v2 is a feature extraction backbone for remote sensing object detection. PKINet-v2 is built upon the modern backbone structures~\cite{liu2022convnet, wang2022pvt, guo2023visual, yu2022metaformer}, and can be incorporated into various oriented object detectors such as Oriented RCNN~\cite{xie2021oriented} to obtain the final object detection results for RSIs.
Each stage $l$ repeats PKINet-v2 Blocks $D_l$ times to refine representation. Each PKINet-v2 Block consists of two residual sub-blocks: the Poly-Kernel Scope (PKS) Block and the Feed-forward Network (FFN) Block. The PKS Block is mainly built upon the PKS Module (\S\ref{sec:PKS}), which synergizes anisotropic strip and isotropic square convolutions to build a multi-scale receptive field for diverse geometries and scales. To enable efficient deployment, HKR Strategy (\S\ref{sec:HKR}) re-parameterizes  the multi branch into a single branch for inference, reducing fragmentation without accuracy loss.

There are four stages arranged sequentially in PKINet-v2. The input and output of stage $l$ are $\bm{F}_{l-1\!}\!\in\!\mathbb{R}^{C_{l-1}\!\times\!H_{l-1}\!\times\!W_{l-1}}$ and $\bm{F}_{l}\!\in\!\mathbb{R}^{C_l\!\times\!H_l\!\times\!W_l}$, respectively. 
Stage $l$ starts with an overlap patch embedding that performs downsampling and channel expansion:
\vspace{-5pt}
\begin{equation}\small
\bm{X}_{l-1} = \texttt{PE}_{l-1}(\bm{F}_{l-1}) \in \mathbb{R}^{C_l\times H_l\times W_l}.
\end{equation}
Then, $l$-th stage repeats PKINet-v2 blocks for $N_l$ times and obtains stage output $\bm{F}_{l} \in \mathbb{R}^{C_l\times H_l\times W_l}$. As shown in Fig.~\ref{fig:framework}, each PKINet-v2 block basically contains a PKS Module (\S\ref{sec:PKS}) and a FFN. The PKS Module builds a multi-scope receptive field while the FFN is used for channel mixing and feature refinement.

\subsection{Poly-Kernel Scope Module}
\label{sec:PKS}
Remote sensing objects exhibit both geometric anisotropy (\textit{e.g.}, bridges) and extreme scale variation (\textit{e.g.}, vehicles vs. courts). 
To jointly handle these properties, we design the Poly-Kernel Scope (PKS) Module, which unifies anisotropic strip kernels and isotropic square kernels and builds a multi-scope receptive field.

As shown in Fig.~\ref{fig:framework}(d), within $n$-th block, given the input feature $\bm{X}_{l-1,n}\!\in\!\mathbb{R}^{C_l\!\times\!H_l\!\times\!W_l}$, PKS Module comprises a small-kernel convolution to grasp local information, followed by a set of parallel depth-wise convolutions, including strip and square-kernel branches with diverse receptive-field scopes, to aggregate multi-scale contextual cues.
\vspace{-3pt}
\begin{equation}\small
\begin{split}
&\bm{L}_{l-1,n} = \texttt{DWConv}_{k_s\!\times k_s}(\bm{X}_{l-1,n}), ~n\!=\!0,\dots,N_l\!-\!1,\\
&\bm{Z}^{(m)}_{l-1,n} = \texttt{DWConv}_{\mathcal{K}^{(m)}}(\bm{L}_{l-1,n}), \quad m=1,\dots,M,
\end{split}
\vspace{-3pt}
\end{equation}
where $M$ is the number of parallel branches, $\bm{L}_{l-1,n}\!\in\!\mathbb{R}^{C_l\!\times\!H_l\!\times\!W_l}$ denotes the local feature extracted by a small $k_s\!\times\!k_s$ depth-wise convolution ($\texttt{DWConv}$), and $\bm{Z}^{(m)}_{l-1,n}\!\in\!\mathbb{R}^{C_l\!\times\!H_l\!\times\!W_l}$ refers to the feature extracted by the $m$-th depth-wise branch with kernel specification $\mathcal{K}^{(m)}$, instantiated by three kernel types:
\vspace{-4pt}
\begin{equation}\small
\texttt{DWConv}_{\mathcal{K}^{(m)}}(\cdot)=
\begin{cases}
\texttt{DWConv}_{k_m\times 1}\!\big(\texttt{DWConv}_{1\times k_m}(\cdot)\big),
& m\in\mathcal{M}_{\text{strip}} \ \text{(axial strip)},\\
\texttt{DWConv}^{d_m}_{k_m\times k_m}(\cdot),\ d_m>1, 
& m\in\mathcal{M}_{\text{sparse}} \ \text{(sparse square)},\\
\texttt{DWConv}^{d_m}_{k_m\times k_m}(\cdot),\ d_m=1, 
& m\in\mathcal{M}_{\text{dense}} \ \text{(dense square)},
\end{cases}
\vspace{-3pt}
\end{equation}
where $\mathcal{M}_{\text{strip}}$, $\mathcal{M}_{\text{sparse}}$, and $\mathcal{M}_{\text{dense}}$ denote the index sets of the axial-strip, sparse-square, and dense-square branches, respectively; $k_m$ is the kernel size of branch $m$, and $d_m$ is its dilation rate (with $d_m>1$ for sparse-square and $d_m=1$ for dense-square). In practice, we set $k_s\!=\!5$ for local feature extraction, and $M=5$ parallel depth-wise branches: one axial-strip branch with $k_m\!=\!19$, three sparse-square branches with $d_m\!=\!3$ and $k_m\in\{7,5,3\}$, and one dense-square branch with $k_m\!=\!3$ to preserve local textures within the aggregation.

Then, the multi-branch features are first normalized by branch-specific Batch Normalization ($\texttt{BN}$) to stabilize training and harmonize different branch responses, and are then fused by a $1\times1$ convolution to characterize the interrelations among different receptive fields and geometries:
\vspace{-4pt}
\begin{equation}\small
\label{eq:pks_output}
\bm{A}_{l-1,n} =
\texttt{Conv}_{1\!\times\!1}\Big(\sum\nolimits_{m=1}^{M}\texttt{BN}^{(m)}(\bm{Z}^{(m)}_{l-1,n})\Big),
\vspace{-3pt}
\end{equation}
where $\bm{A}_{l-1,n}\!\in\!\mathbb{R}^{C_l\!\times\!H_l\!\times\!W_l}$ denotes the spatial attention map.
The $1\!\times\!1$ convolution serves as a channel fusion mechanism to aggregate heterogeneous kernel responses with varying receptive scopes. Similar to~\cite{guo2023visual, guo2022segnext, wang2022pvt}, the final output of the PKS Module is the element-wise product between $\bm{X}_{l-1,n}$ and $\bm{A}_{l-1,n}$:
\vspace{-4pt}
\begin{equation}\small
\label{eq:pks_gate}
\bm{Y}_{l,n} = \bm{X}_{l-1,n}\odot \bm{A}_{l-1,n},
\vspace{-3pt}
\end{equation}
where $\odot$ denotes element-wise multiplication and $\bm{Y}_{l,n}$ is the output of PKS.

\begin{figure*}[t!]
    \centering    \includegraphics[width=0.95\linewidth]{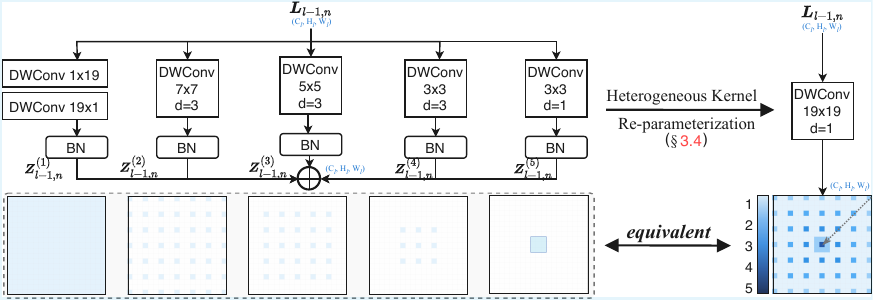}
    \vspace{-8pt}
    \caption{\textbf{Receptive field construction and HKR Strategy.} \textbf{(Left)} The multi-branch PKS Module (\S\ref{sec:PKS}) aggregates heterogeneous depth-wise branches to form a hierarchically densified receptive field with a broad receptive-field range, where global full-span coverage is progressively enriched toward the center. \textbf{(Right)} Heterogeneous Kernel Re-parameterization (HKR, \S\ref{sec:HKR}) algebraically fuses Conv-BN and merges all branches into one equivalent $K_{\max}\!\times\!K_{\max}$ depth-wise convolution, preserving identical outputs while significantly improving inference efficiency. Please refer to \S\ref{sec:method} for details.}
    \vspace{-16pt}
    \label{fig:kernel}
\end{figure*}

As illustrated in Fig.~\ref{fig:kernel}, the PKS Module exhibits a \textit{hierarchically densified} receptive-field pattern, with the following key characteristics:
(i) the axial-strip branch provides global full-span coverage, yielding a gap-free receptive-field domain that is particularly beneficial for slender structures;
(ii) the superposition of multi-scope branches forms an outer-to-inner, progressively densified overlap, where the effective aggregation degree rises from 2 at the boundaries to 5 in center, enabling coarse-to-fine context aggregation across scales. The isotropic, progressively densified aggregation better aligns with square-shaped objects and remains robust to large-scale diversity in RSIs;
(iii) the dense-square branch concentrates the highest coverage multiplicity around the center, thereby strengthening local texture cues and improving sensitivity to small, detail-heavy targets; (iv) PKS Module remains efficient by leveraging low-cost depth-wise axial strip and dilated square kernels to obtain large receptive fields.

\subsection{Heterogeneous Kernel Re-parameterization}
\label{sec:HKR}
Although the design of multi-branch PKS (\S\ref{sec:PKS}) Module is effective, directly executing parallel depth-wise branches during inference leads to fragmented kernel launches and degraded throughput from repeated feature materialization and memory traffic.
To this end, inspired by~\cite{zhang2025scaling}, we propose a Heterogeneous Kernel Re-parameterization (HKR) Strategy, which converts the training-time heterogeneous branches into a single depth-wise large-kernel convolution for deployment, while preserving identical outputs and improving inference efficiency.

\noindent\textbf{Conv-BN Fusion.}
Recall that Eq.~\eqref{eq:pks_output} applies a branch-specific $\texttt{BN}^{(m)}$ on each branch feature $\bm{Z}^{(m)}_{l-1,n}$.
For deployment, we absorb $\texttt{BN}^{(m)}$ into its preceding depth-wise convolution, yielding the following equivalent form:
\vspace{-4pt}
\begin{equation}\small
\label{eq:bn_fuse}
\begin{aligned}
\texttt{BN}^{(m)}(\bm{Z}^{(m)}_{l-1,n})
&=\texttt{BN}^{(m)}\!\Big(\texttt{DWConv}_{\mathcal{K}^{(m)}}(\bm{L}_{l-1,n};\bm{W}^{(m)})\Big) \\
&=\texttt{DWConv}_{\mathcal{K}^{(m)}}(\bm{L}_{l-1,n};\hat{\bm{W}}^{(m)},\hat{\bm{b}}^{(m)}),
\end{aligned}
\vspace{-3pt}
\end{equation}
where $\bm{W}^{(m)}\!\in\!\mathbb{R}^{C_l\times 1\times k_m\times k_m}$ is the depth-wise kernel of branch $m$,
and $(\hat{\bm{W}}^{(m)},\linebreak \hat{\bm{b}}^{(m)})$ are the fused kernel and bias, which can be computed as:
\vspace{-4pt}
\begin{equation}\small
\hat{\bm{W}}^{(m)}=\bm{W}^{(m)}\odot\frac{\bm{\gamma}^{(m)}}{\sqrt{\bm{\sigma}^{2(m)}+\epsilon}},\quad
\hat{\bm{b}}^{(m)}=\bm{\beta}^{(m)}-\bm{\mu}^{(m)}\odot\frac{\bm{\gamma}^{(m)}}{\sqrt{\bm{\sigma}^{2(m)}+\epsilon}},
\vspace{-3pt}
\end{equation}
where $\bm{\gamma}^{(m)}$ and $\bm{\beta}^{(m)}$ are the learned per-channel scale and shift of $\texttt{BN}^{(m)}$,
$\bm{\mu}^{(m)}$ and $\bm{\sigma}^{2(m)}$ are its running mean and running variance,
and $\odot$ denotes channel-wise scaling in depth-wise convolution.

\noindent\textbf{Heterogeneous-to-Homogeneous Kernel Conversion.}
After Conv-BN fusion, all branches in Eq.~\eqref{eq:pks_output} become depth-wise convolutions with parameters $(\hat{\bm{W}}^{(m)},\hat{\bm{b}}^{(m)})$.
For deployment, we merge the $M$ heterogeneous branches into a single depth-wise large-kernel convolution with kernel size $K_{\max}\!\times\!K_{\max}$, where $K_{\max}$ matches the maximum receptive scope.
Concretely, we construct a fused kernel $\bm{W}^{\star}$ and bias $\bm{b}^{\star}\!\in\!\mathbb{R}^{C_l}$ by summing contributions of all branches.
Let $c_0=\big\lfloor K_{\max}/2 \big\rfloor$ denote the center index of the fused kernel.

\emph{(i) Square branches (dense/sparse).}
For a square branch $m$ with kernel size $k_m\times k_m$ and dilation $d_m$ (with $d_m{=}1$ for dense and $d_m{>}1$ for sparse), we place its fused weights into $\bm{W}^{\star}$ by center alignment with dilation-aware offsets:
\vspace{-4pt}
\begin{equation}\small
\label{eq:hkr_square_scatter}
\bm{W}^{\star}\big[:,:,\,c_0+\Delta i,\;c_0+\Delta j\big]\;{+}{=}\;\hat{\bm{W}}^{(m)}[:,:,\,i,\;j],\quad
\bm{b}^{\star}{+}{=}\hat{\bm{b}}^{(m)},
\vspace{-3pt}
\end{equation}
where $i,j\!\in\!\{0,\dots,k_m\!-\!1\}$ and
$\Delta i=(i-\lfloor k_m/2\rfloor)d_m$, $\Delta j=(j-\lfloor k_m/2\rfloor)d_m$.

\emph{(ii) Axial-strip branch.}
For the axial-strip branch, the depth-wise kernel is realized by a $1\times k_m$ convolution followed by a $k_m\times 1$ convolution.
Let their fused kernels be $\hat{\bm{W}}^{(m)}_{1\times k_m}\!\in\!\mathbb{R}^{C_l\times 1\times 1\times k_m}$ and
$\hat{\bm{W}}^{(m)}_{k_m\times 1}\!\in\!\mathbb{R}^{C_l\times 1\times k_m\times 1}$.
We first convert them into an equivalent $k_m\times k_m$ kernel via the channel-wise outer product:
\vspace{-4pt}
\begin{equation}\small
\label{eq:hkr_axial_outer}
\hat{\bm{W}}^{(m)}[:,:,\,i,\;j]
=
\hat{\bm{W}}^{(m)}_{k_m\times 1}[:,:,\,i,\;0]\odot
\hat{\bm{W}}^{(m)}_{1\times k_m}[:,:,\,0,\;j],
\vspace{-3pt}
\end{equation}
where $\odot$ denotes channel-wise multiplication and $i,j\!\in\!\{0,\dots,k_m\!-\!1\}$.
Then, $\hat{\bm{W}}^{(m)}\!\in\!\mathbb{R}^{C_l\times 1\times k_m\times k_m}$ is inserted into the fused kernel $\bm{W}^{\star}$ using Eq.~\eqref{eq:hkr_square_scatter}.

After merging all branches, the summed multi-branch aggregation is equivalent to a single depth-wise convolution:
\vspace{-4pt}
\begin{equation}\small
\label{eq:hkr_equiv_dw}
\sum_{m=1}^{M}\texttt{DWConv}_{\mathcal{K}^{(m)}}(\bm{L}_{l-1,n};\hat{\bm{W}}^{(m)},\hat{\bm{b}}^{(m)})
\equiv
\texttt{DWConv}_{K_{\max}\times K_{\max}}(\bm{L}_{l-1,n};\bm{W}^{\star},\bm{b}^{\star}).
\vspace{-3pt}
\end{equation}
In practice, $K_{\max}{=}19$ and all PKS branches are re-parameterized into a single $19\!\times\!19$ depth-wise convolution after HKR. This conversion collapses parallel kernels into one contiguous operator, which substantially reduces fragmented kernel launches and intermediate feature materialization, leading to higher throughput and lower inference latency while preserving identical outputs.

\begin{table}[t]
  \centering
  \footnotesize
  \setlength{\tabcolsep}{0.6mm}
  \caption{\textbf{Variants of PKINet-v2 backbone.} $C_i$: feature channel number; $D_i$: number of strip blocks in each stage $i$. All metrics are reported with an input resolution of $1024\times1024$, and FPS is tested on a single NVIDIA A100-40G GPU. All variants are built within the framework of Oriented RCNN~\cite{xie2021oriented}. Params and FLOPs are computed for backbones only. Please refer to \S\ref{sec:method} for more details.}
  \vspace{-10pt}

  \resizebox{0.98\linewidth}{!}{%
  \begin{tabular}{c||cc|cc|cc} \hline
  \rowcolor[rgb]{0.92,0.92,0.92}
  Variant & \{{$C_1$, $C_2$, $C_3$, $C_4$}\} & \{{$D_1$, $D_2$, $D_3$, $D_4$}\} & \#P & FLOPs & FPS & FPS-HKR \\ \hline \hline
  $\star$ PKINet-v2-T & \{32, 64, 160, 256\} & \{3, 3, 5, 2\} & 4.0M & 18.2G & 46.2 & 58.0 \\
  $\star$ PKINet-v2-S & \{64, 128, 320, 512\} & \{2, 2, 4, 2\} & 13.6M & 54.0G & 43.4 & 54.6 \\ \hline
  \end{tabular}%
  }

  \label{tab:variants}
  \vspace{-16pt}
\end{table}

\subsection{Implementation Details}
\label{sec:imp_details_v2}
We present two variants of our backbone, namely \textbf{PKINet-v2-T} (Tiny) and \textbf{PKINet-v2-S} (Small).
PKINet-v2 follows a four-stage hierarchical design. Given an input of size $H\times W$,
the $l$-th stage output has spatial resolution $H_l{=}H/2^{(l+1)}$ and $W_l{=}W/2^{(l+1)}$ for $l{=}1,\dots,4$.
Each stage starts with an overlap patch embedding for downsampling and channel expansion, followed by $D_l$ PKINet-v2 Blocks.
As summarized in Tab.~\ref{tab:variants}, \textbf{PKINet-v2-T} uses
$\{C_1,\allowbreak C_2,\allowbreak C_3,\allowbreak C_4\}\allowbreak=\{32,\allowbreak 64,\allowbreak 160,\allowbreak 256\}$
and
$\{D_1,\allowbreak D_2,\allowbreak D_3,\allowbreak D_4\}\allowbreak=\{3,\allowbreak 3,\allowbreak 5,\allowbreak 2\}$,
while \textbf{PKINet-v2-S} uses
$\{C_1,\allowbreak C_2,\allowbreak C_3,\allowbreak C_4\}\allowbreak=\{64,\allowbreak 128,\allowbreak 320,\allowbreak 512\}$
and
$\{D_1,\allowbreak D_2,\allowbreak D_3,\allowbreak D_4\}\allowbreak=\{2,\allowbreak 2,\allowbreak 4,\allowbreak 2\}$.

\section{Experiment}
\subsection{Experimental Setup}
\noindent\textbf{Datasets.} We conduct extensive experiments on four popular remote sensing object detection datasets:
\begin{itemize}
    \item \textbf{DOTA-v1.0} \cite{xia2018dota} is a$_{\!}$ large-scale$_{\!}$ dataset$_{\!}$ for$_{\!}$ remote$_{\!}$ sensing$_{\!}$ detection$_{\!}$, which$_{\!}$ contains$_{\!}$ 2806$_{\!}$ images, 188,282 instances, and 15 categories with a large variety of orientations and scales.
    The dataset comprises 1,411, 458, and 937 images for \texttt{train}, \texttt{val}, and \texttt{test}, respectively.
    \item \textbf{DOTA-v1.5} \cite{xia2018dota} is a more challenging dataset based on DOTA-v1.0 which is released for DOAI Challenge 2019. This iteration includes the addition of a novel category named \texttt{Container Crane} (CC) and a substantial increase in the number of minuscule instances that are less than 10 pixels, containing 403,318 instances in total.
    \item \textbf{HRSC2016} \cite{liu2017high} is a remote sensing dataset for ship detection that contains 1061 aerial images whose sizes range from $300 \times 300$ to $1500 \times 900$. The images are split into 436/181/444 for \texttt{train/val/test}.
    \item \textbf{DIOR-R} \cite{cheng2022anchor} provides OBB annotations based on DIOR \cite{li2020object} dataset. It contains 23,463 images with the size of $800 \times 800$ and 192,518 annotations.
\end{itemize}

\begin{table*}[t]
 \vspace{-5pt}
 \caption{\textbf{Experimental results on DOTA-v1.0 dataset} \cite{xia2018dota} under single-scale training and testing setting. PKINet-v2-S is pretrained on ImageNet-1K \cite{deng2009imagenet} for 300 epochs similar to previous methods \cite{Li_2023_ICCV,cai2024poly,yuan2025strip,lu2025lwganet}. $^\dag$: Use the Strip Head~\cite{yuan2025strip}. See \S\ref{sec:quantitative_results}.}
 \vspace{-10pt}
\setlength{\tabcolsep}{2pt}
\renewcommand\arraystretch{1.15}
\scriptsize      
\centering
\resizebox{\textwidth}{!}{
\hspace{-1.0em}
\begin{tabular}{r|c||c|c|ccccccccccccccc} 
\thickhline
\rowcolor[rgb]{0.92,0.92,0.92} \textbf{Method}       & \textbf{mAP} $\uparrow$  & \#\textbf{P} $\downarrow$ & \textbf{FLOPs} $\downarrow$    & PL    & BD    & BR    & GTF   & SV    & LV    & SH    & TC    & BC    & ST    & SBF   & RA    & HA    & SP    & HC  \\ 
\hline
\hline
 \multicolumn{2}{l}{\textbf{\textit{DETR-based}}}\\
\hline
 AO\textsuperscript{2}-DETR~\cite{dai2022ao2}  & 70.91 & 40.8M & 304G & 87.99 & 79.46 & 45.74 & 66.64 & 78.90 & 73.90 & 73.30 & 90.40 & 80.55 & 85.89 & 55.19 & 63.62 & 51.83 & 70.15 & 60.04  \\ 
 O\textsuperscript{2}-DETR~\cite{ma2021oriented}  & 72.15 & - & - & 86.01 & 75.92 & 46.02 & 66.65 & 79.70 & 79.93 & 89.17 & 90.44 & 81.19 & 76.00 & 56.91 & 62.45 & 64.22 & 65.80 & 58.96  \\ 
 ARS-DETR~\cite{zeng2023ars}  & 73.79 & 41.6M  & & 86.61 & 77.26 & 48.84 & 66.76 & 78.38 & 78.96 & 87.40 & 90.61 & 82.76 & 82.19 & 54.02 & 62.61 & 72.64 & 72.80 & 64.96  \\
\hline
\hline
\multicolumn{2}{l}{\textbf{\textit{One-stage}}}\\
\hline
SASM~\cite{hou2022shape}  & 74.92 & 36.6M & 194G & 86.42 & 78.97 & 52.47 & 69.84 & 77.30 & 75.99 & 86.72 & 90.89 & 82.63 & 85.66 & 60.13 & 68.25 & 73.98 & 72.22 & 62.37  \\ 
R3Det-GWD~\cite{yang2021rethinking} & 76.34 & 41.9M & 336G & 88.82 & 82.94 & 55.63 & 72.75 & 78.52 & 83.10 & 87.46 & 90.21 & 86.36 & 85.44 & 64.70 & 61.41 & 73.46 & 76.94 & 57.38  \\ 
R3Det-KLD~\cite{yang2021learning} & 77.36 & 41.9M & 336G & 88.90 & 84.17 & 55.80 & 69.35 & 78.72 & 84.08 & 87.00 & 89.75 & 84.32 & 85.73 & 64.74 & 61.80 & 76.62 & 78.49 & 70.89  \\ 
O-RepPoints~\cite{li2022oriented}  & 75.97 & 36.6M & 194G & 87.02 & 83.17 & 54.13 & 71.16 & 80.18 & 78.40 & 87.28 & 90.90 & 85.97 & 86.25 & 59.90 & 70.49 & 73.53 & 72.27 & 58.97  \\ 
Rotated FCOS~\cite{tian2019fcos} & 72.45 & 31.9M & 207G & 88.52 & 77.54 & 47.06 & 63.78 & 80.42 & 80.50 & 87.34 & 90.39 & 77.83 & 84.13 & 55.45 & 65.84 & 66.02 & 72.77 & 49.17  \\
R3Det~\cite{yang2021r3det} & 69.70 & 41.9M & 336G & 89.00 & 75.60 & 46.64 & 67.09 & 76.18 & 73.40 & 79.02 & 90.88 & 78.62 & 84.88 & 59.00 & 61.16 & 63.65 & 62.39 & 37.94  \\
S$^2$ANet~\cite{han2021align} & 74.12 & 38.5M & 198G & 89.11 & 82.84 & 48.37 & 71.11 & 78.11 & 78.39 & 87.25 & 90.83 & 84.90 & 85.64 & 60.36 & 62.60 & 65.26 & 69.13 & 57.94  \\
\hline
\hline
\multicolumn{2}{l}{\textbf{\textit{Two-stage}}}\\
\hline
SCRDet~\cite{yang2019scrdet} & 72.61 & 41.9M & - & 89.98 & 80.65 & 52.09 & 68.36 & 68.36 & 60.32 & 72.41 & 90.85 & 87.94 & 86.86 & 65.02 & 66.68 & 66.25 & 68.24 & 65.21  \\ 
G.V.~\cite{xu2020gliding}  & 75.02 & 41.1M & 198G & 89.64 & 85.00 & 52.26 & 77.34 & 73.01 & 73.14 & 86.82 & 90.74 & 79.02 & 86.81 & 59.55 & 70.91 & 72.94 & 70.86 & 57.32  \\ 
CenterMap~\cite{long2021creating} & 71.59 & 41.1M & 198G & 89.02 & 80.56 & 49.41 & 61.98 & 77.99 & 74.19 & 83.74 & 89.44 & 78.01 & 83.52 & 47.64 & 65.93 & 63.68 & 67.07 & 61.59  \\ 
ReDet~\cite{han2021redet}  & 76.25 & 31.6M & - & 88.79 & 82.64 & 53.97 & 74.00 & 78.13 & 84.06 & 88.04 & 90.89 & 87.78 & 85.75 & 61.76 & 60.39 & 75.96 & 68.07 & 63.59  \\ 
RoI Trans \cite{ding2019learning} & 74.05 & 55.1M & 200G & 89.01 & 77.48 & 51.64 & 72.07 & 74.43 & 77.55 & 87.76 & 90.81 & 79.71 & 85.27 & 58.36 & 64.11 & 76.50 & 71.99 &  54.06  \\
R Faster RCNN ~\cite{ren2015faster} & 73.17 & 41.1M & 211G & 89.40 & 81.81 & 47.28 & 67.44 & 73.96 & 73.12 & 85.03 & 90.90 & 85.15 & 84.90 & 56.60 & 64.77 & 64.70 & 70.28 & 62.22  \\
O-RCNN~\cite{xie2021oriented} & 75.87 & 41.1M & 199G & 89.46 & 82.12 & 54.78 & 70.86 & 78.93 & 83.00 & 88.20 & 90.90 & 87.50 & 84.68 & 63.97 & 67.69 & 74.94 & 68.84 & 52.28  \\
ARC~\cite{pu2023adaptive} & 77.35 & 74.4M & 217G & 89.40 & 82.48 & 55.33 & 73.88 & 79.37 & 84.05 & 88.06 & 90.90 & 86.44 & 84.83 & 63.63 & 70.32 & 74.29 & 71.91 &  65.43  \\
LSKNet-S~\cite{Li_2023_ICCV}  & 77.49 & 31.0M  & 161G & 89.66 & 85.52 & 57.72 & 75.70 & 74.95 & 78.69 & 88.24 & 90.88 & 86.79 & 86.38 & 66.92 & 63.77 & 77.77 & 74.47 & 64.82     \\
Strip RCNN-S$^\dag$~\cite{yuan2025strip} & 80.06 & 31.2M & 204G & 88.91 & 86.38 & 57.44 & 76.37 & 79.73 & 84.38 & 88.25 & 90.86 & 86.71 & 87.45 & 69.89 & 66.82 & 79.25 & 82.91 & 75.58 \\
LWGANet-L2~\cite{lu2025lwganet} & 79.02 & 29.2M & 159G & 89.49 & 85.48 & 54.93 & 77.12 & 81.59 & 85.64 & 88.43 & 90.85 & 87.23 & 86.78 & 67.47 & 65.06 & 78.23 & 73.33 & 73.66 \\
PKINet-v1-S~\cite{cai2024poly} & 78.39 & 30.8M & 184G & 89.72 & 84.20 & 55.81 & 77.63 & 80.25 & 84.45 & 88.12 & 90.88 & 87.57 & 86.07 & 66.86 & 70.23 & 77.47 & 73.62 & 62.94  \\
\rowcolor[rgb]{0.92,0.92,0.92} \textbf{PKINet-v2-T} ~~~ & 79.36 & 21.0M & 137G & 89.74 & 86.10 & 57.26 & 78.97 & 80.37 & 84.95 & 88.48 & 90.88 & 87.80 & 86.49 & 67.69 & 69.60 & 77.31 & 73.92 & 70.93\\ 
\rowcolor[rgb]{0.92,0.92,0.92} \textbf{PKINet-v2-S} ~~~ & 80.46 & 30.7M & 173G & 89.53 & 85.19 & 57.72 & 78.63 & 81.21 & 86.38 & 88.18 & 90.84 & 87.95 & 86.25 & 71.97 & 67.76 & 78.15 & 81.80 & 75.34\\ 
\rowcolor[rgb]{0.92,0.92,0.92} \textbf{PKINet-v2-S$^\dag$} ~~ & \textbf{80.68} & 31.4M & 206G & 89.81 & 86.05 & 60.82 & 79.62 & 81.51 & 86.09 & 88.34 & 90.95 & 88.53 & 87.09 & 69.08 & 69.95 & 78.49 & 80.60 & 73.37\\ 
\hline
\end{tabular}}
\label{tab:var_arch}
\vspace{-8pt}
\end{table*}

\begin{table*}[t]
\centering
\caption{\textbf{Experimental results on DOTA-v1.5 dataset} \cite{xia2018dota} compared with state-of-the-art methods with single-scale training and testing. PKINet-v2-S backbone is pretrained on ImageNet-1K \cite{deng2009imagenet} for 300 epochs, as the compared methods \cite{han2021redet, xu2023dynamic, cai2024poly, yuan2025strip}. PKINet-v2-S is built within the framework of Oriented RCNN \cite{xie2021oriented}. See \S\ref{sec:quantitative_results} for details.}
\vspace{-10pt}
\setlength\tabcolsep{5pt}
\renewcommand\arraystretch{1.15}
\resizebox{\textwidth}{!}{
\begin{tabular}{r||c|cccccccccccccccc} 
\spthickhline
\rowcolor[rgb]{0.92,0.92,0.92} \textbf{Method} & \textbf{mAP $\uparrow$} & PL & BD & BR & GTF & SV & LV & SH & TC & BC & ST & SBF & RA & HA & SP & HC & CC \\ 
\hline
\hline
RetinaNet-O~\cite{lin2017focal} & 59.16 & 71.43 & 77.64 & 42.12 & 64.65 & 44.53 & 56.79 & 73.31 & 90.84 & 76.02 & 59.96 & 46.95 & 69.24 & 59.65 & 64.52 & 48.06 & 0.83 \\
FR-O~\cite{ren2015faster}       & 62.00 & 71.89 & 74.47 & 44.45 & 59.87 & 51.28 & 68.98 & 79.37 & 90.78 & 77.38 & 67.50 & 47.75 & 69.72 & 61.22 & 65.28 & 60.47 & 1.54 \\
Mask RCNN~\cite{he2017mask}    & 62.67 & 76.84 & 73.51 & 49.90 & 57.80 & 51.31 & 71.34 & 79.75 & 90.46 & 74.21 & 66.07 & 46.21 & 70.61 & 63.07 & 64.46 & 57.81 & 9.42 \\
HTC~\cite{chen2019hybrid}       & 63.40 & 77.80 & 73.67 & 51.40 & 63.99 & 51.54 & 73.31 & 80.31 & 90.48 & 75.12 & 67.34 & 48.51 & 70.63 & 64.84 & 64.48 & 55.87 & 5.15 \\
ReDet~\cite{han2021redet}       & 66.86 & 79.20 & 82.81 & 51.92 & 71.41 & 54.08 & 75.73 & 80.92 & 90.83 & 75.81 & 68.64 & 49.29 & 72.03 & 73.36 & 70.55 & 63.33 & 11.53 \\
DCFL~\cite{xu2023dynamic}       & 67.37 & - & - & - & - & 56.72 & - & 80.87 & - & - & 75.65 & - & - & - & - & - & - \\
LSKNet-S~\cite{Li_2023_ICCV}    & 70.26 & 72.05 & 84.94 & 55.41 & 74.93 & 52.42 & 77.45 & 81.17 & 90.85 & 79.44 & 69.00 & 62.10 & 73.72 & 77.49 & 75.29 & 55.81 & 42.19 \\
Strip RCNN-S~\cite{yuan2025strip} & 72.27 & 80.04 & 83.26 & 54.40 & 75.38 & 52.46 & 81.44 & 88.53 & 90.83 & 84.80 & 69.65 & 65.93 & 73.28 & 74.61 & 74.04 & 69.70 & 38.98 \\
PKINet-v1-S~\cite{cai2024poly}  & 71.47 & 80.31 & 85.00 & 55.61 & 74.38 & 52.41 & 76.85 & 88.38 & 90.87 & 79.04 & 68.78 & 67.47 & 72.45 & 76.24 & 74.53 & 64.07 & 37.13 \\
\hline
\hline
\rowcolor[rgb]{0.92,0.92,0.92} \textbf{PKINet-v2-S} ~~~  & \textbf{73.57} & 80.40 & 85.15 & 56.33 & 75.14 & 52.70 & 82.69 & 89.12 & 90.89 & 79.53 & 68.96 & 65.73 & 73.72 & 77.76 & 76.00 & 72.79 & 50.15 \\
\hline
\end{tabular}
}
\label{tab:main_performance_dotav15}
\vspace{-20pt}
\end{table*}

\begin{table*}[t]
\centering
\vspace{-8pt}

\begin{minipage}[t]{0.59\textwidth}
\centering
\captionof{table}{\textbf{Experimental results on HRSC2016 dataset} \cite{liu2017high}. PKINet-S is pretrained on ImageNet-1K  \cite{deng2009imagenet} for 300 epochs which is consistent with previous methods \cite{Li_2023_ICCV, cai2024poly, yuan2025strip, lu2025lwganet} and built within the framework of Oriented RCNN \cite{xie2021oriented}. mAP (07/12): VOC 2007 \cite{voc2007}/2012 \cite{voc2012} metrics. Refer to \S\ref{sec:quantitative_results}.}
\label{tab:performance_hrsc}
\vspace{-8pt}

\setlength\tabcolsep{8pt}
\renewcommand\arraystretch{1.11}
\resizebox{0.99\linewidth}{!}{
\begin{tabular}{r||c|c|c|c}
\spthickhline
\rowcolor[rgb]{0.92,0.92,0.92}
\textbf{Method}~~~~~ & \textbf{mAP(07)}$\uparrow$ & \textbf{mAP(12)}$\uparrow$ & \#\textbf{P}$\downarrow$ & \textbf{FLOPs}$\downarrow$ \\
\hline
\hline
DRN~\cite{pan2020dynamic}              & -     & 92.70 & -     & -    \\
DAL~\cite{ming2021dynamic}            & 89.77 & -     & 36.4M & 216G \\
GWD~\cite{yang2021rethinking}         & 89.85 & 97.37 & 47.4M & 456G \\
RoI Trans.~\cite{ding2019learning}    & 86.20 & -     & 55.1M & 200G \\
G.V.~\cite{xu2020gliding}             & 88.20 & -     & 41.1M & -    \\
CenterMap~\cite{long2021creating}     & -     & 92.80 & 41.1M & 198G \\
AOPG~\cite{cheng2022anchor}           & 90.34 & 96.22 & -     & -    \\
R3Det~\cite{yang2021r3det}            & 89.26 & 96.01 & 41.9M & 336G \\
S$^2$ANet~\cite{han2021align}         & 90.17 & 95.01 & 38.6M & 198G \\
ReDet~\cite{han2021redet}             & 90.46 & 97.63 & 31.6M & -    \\
O-RCNN~\cite{xie2021oriented}         & 90.50 & 97.60 & 41.1M & 199G \\
RTMDet~\cite{lyu2022rtmdet}           & 90.60 & 97.10 & 54.0M & 205G \\
O-RepPoints~\cite{li2022oriented}     & 90.38 & 97.26 & 36.6M & 194G \\
LSKNet-S~\cite{Li_2023_ICCV}          & 90.65 & 98.46 & 31.0M & 161G \\
Strip RCNN-S~\cite{yuan2025strip}     & 90.60 & 98.70 & 30.5M & 159G \\
PKINet-v1-S~\cite{cai2024poly}        & 90.70 & 98.54 & 30.8M & 184G \\ \hline\hline
\rowcolor[rgb]{0.92,0.92,0.92} \textbf{PKINet-v2-S}~~~    & \textbf{90.75} & \textbf{98.84} & 30.7M & 173G \\
\hline
\end{tabular}}
\vspace{-4pt}
\end{minipage}
\hfill
\begin{minipage}[t]{0.39\textwidth}
\centering
\captionof{table}{\textbf{Experimental results on DIOR-R dataset}~\cite{li2020object}. Following previous methods \cite{Li_2023_ICCV, cai2024poly, yuan2025strip, lu2025lwganet}, PKINet-v2-S is pretrained on ImageNet-1K \cite{deng2009imagenet} for 300 epochs and built within the framework of Oriented RCNN \cite{xie2021oriented}. See \S\ref{sec:quantitative_results}.}
\label{tab:performance_dior}
\vspace{-6pt}

\setlength{\tabcolsep}{3pt}
\renewcommand\arraystretch{1.15}
\resizebox{\linewidth}{!}{
\begin{tabular}{r||c|c|c}
\hline
\rowcolor[rgb]{0.92,0.92,0.92}\textbf{Method}~~~~~~ & \textbf{mAP} $\uparrow$ & \#\textbf{P} $\downarrow$ & \textbf{FLOPs} $\downarrow$ \\
\hline
\hline
RetinaNet-O~\cite{lin2017focal} & 57.55 & - & - \\
FR-O~\cite{ren2015faster} & 59.54 & 41.1M & 198G \\
TIOE-Det~\cite{ming2023task} & 61.98 & - & - \\
RT~\cite{ding2019learning} & 63.87 & - & - \\
O-RCNN~\cite{xie2021oriented} & 64.30 & 41.1M & 199G \\
GGHL~\cite{huang2022general} & 66.48 & - & - \\
ARS-DETR~\cite{zeng2023ars} & 66.12 & 41.6M & - \\
DCFL~\cite{xu2023dynamic} & 66.80 & - & - \\
Oriented Rep~\cite{li2022oriented} & 66.71 & 36.6M & - \\
LSKNet-S~\cite{Li_2023_ICCV} & 65.90 & 31.0M & 161G \\
Strip RCNN-S~\cite{yuan2025strip} & 68.70 & 31.2M & 204G \\
LWGANet-L2~\cite{lu2025lwganet} & 68.53 & 29.2M & 159G \\
PKINet-v1-S~\cite{cai2024poly} & 67.03 & 30.8M & 184G \\ \hline\hline
\rowcolor[rgb]{0.92,0.92,0.92} \textbf{PKINet-v2-S}~~~ & \textbf{69.40} & 30.7M & 173G \\
\hline
\end{tabular}}
\vspace{-10pt}
\end{minipage}

\vspace{-20pt}
\end{table*}

\noindent\textbf{Training.} Our training process consists of ImageNet \cite{deng2009imagenet} pretraining and remote sensing object detector training. For ImageNet pretraining, PKINet-v2 is trained on the ImageNet-1K. In the main experiment, we train it for 300 epochs for higher performance like previous works \cite{xie2021oriented,yang2021r3det,pu2023adaptive, Li_2023_ICCV,cai2024poly, yuan2025strip}. For ablation studies, we train for only 100 epochs for faster comparisons. We use 4 GPUs with a total batch size of 1024 for pretraining. For remote sensing object detector training, experiments are conducted in MMRotate \cite{zhou2022mmrotate} framework. To compare with other methods, we use \texttt{trainval sets} of these benchmarks and their \texttt{test sets} for testing. Following the settings of previous methods \cite{han2021redet, xie2021oriented, yang2021r3det, zeng2023ars}, we crop original images into $1024 \times 1024$ patches with overlaps of 200 for DOTA-v1.0 and DOTA-v1.5 datasets. For HRSC2016 and DIOR-R datasets, the input size is set as $800 \times 800$. Models are trained for 36 epochs for all datasets. We employ AdamW \cite{kingma2014adam} optimizer with a weight decay of 0.05. The initial learning rate is set to 0.0001. All FLOPs and FPS reported are calculated when the input image size is $1024 \times 1024$. FPS is tested on a single NVIDIA A100-40G GPU. Images undergo random resizing and flipping during training following previous methods \cite{han2021redet, xie2021oriented, yang2021r3det, zeng2023ars}. Five-run average mAP of our method is reported for HRSC2016 and DIOR-R.

\noindent\textbf{Testing.} The image resolution at the testing stage remains consistent with the training stage. For fairness, we do not apply any test-time data augmentation.

\noindent\textbf{Evaluation$_{\!}$ Metric.$_{\!}$} The$_{\!}$ mean$_{\!}$ average$_{\!}$ precision$_{\!}$ (mAP)$_{\!}$ and$_{\!}$ the Average Precision at an IoU threshold of 0.5 ($\mathrm{AP}_{50}$) are reported.

\noindent\textbf{Reproducibility.$_{\!}$} Our$_{\!}$ algorithm$_{\!}$ is$_{\!}$ implemented$_{\!}$ in$_{\!}$ PyTorch. We use 4 NVIDIA A100-40G GPUs for ImageNet pretraining and downstream training and testing. 

\vspace{-6pt}
\subsection{Quantitative Results}
\label{sec:quantitative_results}
\noindent\textbf{Performance on DOTA-v1.0} \cite{xia2018dota}. To demonstrate the effectiveness of our proposed method, we compare PKINet-v2 with state-of-the-art remote sensing backbones. As reported in Tab.~\ref{tab:fps_comparison} and \ref{tab:var_arch}, when equipped with the advanced Oriented RCNN~\cite{xie2021oriented}, PKINet-v2-S achieves a remarkable mAP of \textbf{80.46\%}, surpassing Strip RCNN-S~\cite{yuan2025strip} and PKINet-v1-S~\cite{cai2024poly}, by \textbf{0.4\%} and \textbf{2.07\%}, respectively.

Notably, PKINet-v2 effectively addresses the geometric and spatial complexity challenges.
Regarding \textbf{geometric complexity}, previous isotropic backbones like PKINet-v1-S struggle with high-aspect-ratio targets, achieving only 55.81\% on the Bridge (BR) category. By integrating anisotropic kernels, PKINet-v2-S boosts this to \textbf{57.72\%}, surpassing even the geometry-specialized Strip RCNN-S.
\begin{wraptable}{r}{0.5\textwidth}
  \vspace{-12pt}
  \centering
  \caption{\textbf{Experimental results across different detection frameworks on DOTA-v1.0 dataset}~\cite{xia2018dota}. Refer to \S\ref{sec:quantitative_results}.}
  \label{tab:stripnet_replace}

  \setlength{\tabcolsep}{3pt}
  \renewcommand\arraystretch{1.15}

  \resizebox{\linewidth}{!}{%
  \begin{tabular}{r|r||c|c|c}
    \hline
    \rowcolor[rgb]{0.92,0.92,0.92}
    \textbf{Method}~~~ & \textbf{Backbone}~~ & \textbf{mAP} $\uparrow$ & \textbf{\#P} $\downarrow$ & \textbf{FLOPs} $\downarrow$ \\
    \hline
    \hline

    \multirow{3}{*}{\parbox{2cm}{\raggedleft Rotated~~~~~~\\FCOS~\cite{tian2019fcos}}}
    & ResNet-50~\cite{he2016deep} & 72.45 & 41.1M & 211.4G \\
    \cline{2-5}
    & PKINet-v1~\cite{cai2024poly} & 74.86 & 13.7M & 70.2G \\
    \cline{2-5}
    & \cellcolor[rgb]{0.92,0.92,0.92} PKINet-v2~~~~
      & \cellcolor[rgb]{0.92,0.92,0.92}\textbf{77.13}
      & \cellcolor[rgb]{0.92,0.92,0.92}13.6M
      & \cellcolor[rgb]{0.92,0.92,0.92}54.0G \\
    \hline

    \multirow{4}{*}{R3Det~\cite{yang2021r3det}}
    & ResNet-50~\cite{he2016deep} & 69.70 & 41.1M & 211.4G \\
    \cline{2-5}
    & ARC-R50~\cite{pu2023adaptive} & 72.32 & 56.5M & 86.6G \\ \cline{2-5}
    & PKINet-v1-S~\cite{cai2024poly} & 75.89 & 13.7M & 70.2G \\
    \cline{2-5}
    & \cellcolor[rgb]{0.92,0.92,0.92} PKINet-v2-S~~~~
      & \cellcolor[rgb]{0.92,0.92,0.92}\textbf{77.64}
      & \cellcolor[rgb]{0.92,0.92,0.92}13.6M
      & \cellcolor[rgb]{0.92,0.92,0.92}54.0G \\
    \hline

    \multirow{4}{*}{S$^2$ANet~\cite{han2021align}}
    & ResNet-50~\cite{he2016deep} & 74.12 & 41.1M & 211.4G \\
    \cline{2-5}
    & ARC-R50~\cite{pu2023adaptive} & 75.49 & 56.5M & 86.6G \\ \cline{2-5}
    & PKINet-v1-S~\cite{cai2024poly} & 77.83 & 13.7M & 70.2G \\
    \cline{2-5}
    & \cellcolor[rgb]{0.92,0.92,0.92} PKINet-v2-S~~~~
      & \cellcolor[rgb]{0.92,0.92,0.92}\textbf{79.68}
      & \cellcolor[rgb]{0.92,0.92,0.92}13.6M
      & \cellcolor[rgb]{0.92,0.92,0.92}54.0G \\
    \hline

    \multirow{3}{*}{RoI Trans.~\cite{ding2019learning}}
    & ResNet-50~\cite{he2016deep} & 74.05 & 41.1M & 211.4G \\
    \cline{2-5}
    & PKINet-v1-S~\cite{cai2024poly} & 77.17 & 13.7M & 70.2G \\
    \cline{2-5}
    & \cellcolor[rgb]{0.92,0.92,0.92} PKINet-v2-S~~~~
      & \cellcolor[rgb]{0.92,0.92,0.92}\textbf{79.87}
      & \cellcolor[rgb]{0.92,0.92,0.92}13.6M
      & \cellcolor[rgb]{0.92,0.92,0.92}54.0G \\
    \hline

    \multirow{4}{*}{\parbox{2cm}{\raggedleft Rotated~~~~~~\\Faster~~~~~~~\\RCNN~\cite{ren2015faster}}}
    & ResNet-50~\cite{he2016deep} & 73.17 & 41.1M & 211.4G \\
    \cline{2-5}
    & ARC-R50~\cite{pu2023adaptive} & 74.75 & 56.5M & 86.6G \\ \cline{2-5}
    & PKINet-v1-S~\cite{cai2024poly} & 76.45 & 13.7M & 70.2G \\
    \cline{2-5}
    & \cellcolor[rgb]{0.92,0.92,0.92} PKINet-v2-S~~~~
      & \cellcolor[rgb]{0.92,0.92,0.92}\textbf{79.02}
      & \cellcolor[rgb]{0.92,0.92,0.92}13.6M
      & \cellcolor[rgb]{0.92,0.92,0.92}54.0G \\
    \hline

    \multirow{7}{*}{O-RCNN~\cite{xie2021oriented}}
    & ResNet-50~\cite{he2016deep} & 75.87 & 41.1M & 211.4G \\
    \cline{2-5}
    & ARC-R50~\cite{pu2023adaptive} & 77.35 & 56.5M & 86.6G \\ \cline{2-5}
    & LSKNet-S~\cite{Li_2023_ICCV} & 77.49 & 14.4M & 54.4G \\ \cline{2-5}
    & StripNet-S~\cite{yuan2025strip} & 79.85 & 13.3M & 52.3G \\ \cline{2-5}
    & LWGANet-L2~\cite{lu2025lwganet} & 79.02 &12.0M & 38.8G \\ \cline{2-5}
    & PKINet-v1-S~\cite{cai2024poly} & 78.39 & 13.7M & 70.2G \\
    \cline{2-5}
    & \cellcolor[rgb]{0.92,0.92,0.92} PKINet-v2-S~~~~
      & \cellcolor[rgb]{0.92,0.92,0.92}\textbf{80.46}
      &\cellcolor[rgb]{0.92,0.92,0.92}13.6M
      &\cellcolor[rgb]{0.92,0.92,0.92}54.0G \\
    \hline

  \end{tabular}%
  }

  \vspace{-26pt}
\end{wraptable} 
This validates that our hybrid kernel design successfully captures slender geometries.
Regarding \textbf{spatial complexity}, PKINet-v2-S demonstrates superior robustness across the spectrum. 
For small instances like Small Vehicle (SV), it achieves \textbf{81.21\%}, significantly surpassing LSKNet-S.
Simultaneously, it effectively handles large-scale targets like Soccer Ball Field (SBF), reaching \textbf{71.97\%} and outperforming Strip RCNN-S (69.89\%). 
This confirms that our multi-scope design successfully accommodates extreme scale variations.

Furthermore, to validate the robustness, we integrate PKINet-v2 into various detection frameworks as shown in Table \ref{tab:stripnet_replace}. 
PKINet-v2 brings substantial improvements compared to the ResNet-50 baseline: \textbf{4.68\%} for Rotated FCOS~\cite{tian2019fcos}, \textbf{+7.94\%} for R3Det~\cite{yang2021r3det}, \textbf{+5.56\%} for S$^2$ANet~\cite{han2021align}, \textbf{+5.82\%} for RoI Transformer~\cite{ding2019learning}, \textbf{5.85\%} for Rotated Faster RCNN~\cite{ren2015faster}, and \textbf{4.59\%} for Oriented RCNN~\cite{xie2021oriented}.

\noindent\textbf{Performance on DOTA-v1.5} \cite{xia2018dota}. DOTA-v1.5 is more challenging due to the additional \texttt{CC} category and the increased number of tiny instances. As shown in Table \ref{tab:main_performance_dotav15}, PKINet-v2-S achieves an outstanding mAP of \textbf{73.57\%}. This performance outperforms the previous state-of-the-art method Strip RCNN-S~\cite{yuan2025strip} by \textbf{1.3\%} and surpasses PKINet-v1-S~\cite{cai2024poly} by \textbf{2.1\%}.
This indicates strong tiny-detail preservation of PKINet-v2.

\noindent\textbf{Performance on HRSC2016} \cite{liu2017high}. As shown in Table \ref{tab:performance_hrsc}, PKINet-v2-S sets a new state-of-the-art on HRSC2016 with \textbf{90.75\%} (07) and \textbf{98.84\%} (12) mAP, demonstrating superior capability in detecting slender ships.

\noindent\textbf{Performance on DIOR-R} \cite{cheng2022anchor}. As shown in Tab.~\ref{tab:performance_dior}, PKINet-v2-S achieves the best performance of \textbf{69.40\%}, outperforming LSKNet-S~\cite{Li_2023_ICCV} by a significant margin of \textbf{3.5\%} and exceeding Strip RCNN-S~\cite{yuan2025strip} by \textbf{0.7\%}.

\noindent\textbf{Development Efficiency}. As evidenced in Tables \ref{tab:fps_comparison} and \ref{tab:variants}, leveraging the proposed HKR Strategy, PKINet-v2-S effectively eliminates the memory fragmentation inherent in multi-branch architectures. It achieves a remarkable inference speed of \textbf{54.6 FPS}, representing a \textbf{3.9$\times$} acceleration over PKINet-v1-S and surpassing other state-of-the-art remote sensing backbones.

\noindent\textbf{Performance Analysis}.
Tab.~\ref{tab:dota_ratio_size_tab} summarizes the mAP across aspect-ratio bins and object-size bins on DOTA-v1.0~\cite{xia2018dota}.
PKINet-v2-S consistently achieves higher mAP in all bins, indicating improved robustness to both geometric anisotropy and scale variation in RSIs.

\begin{table*}[h]
\centering
\vspace{-16pt}
\caption{\textbf{Performance analysis on DOTA-v1.0}~\cite{xia2018dota}.
\textbf{(Left)} mAP across object aspect-ratio bins.
\textbf{(Right)} mAP across object-size bins (relative area). All models are trained on the \texttt{train} set and evaluated on the \texttt{val} set. Please refer to \S\ref{sec:quantitative_results} for details.
}
\label{tab:dota_ratio_size_tab}
\vspace{-10pt}

\begin{subtable}{0.496\linewidth}
\centering
\setlength{\tabcolsep}{3pt}
\renewcommand\arraystretch{1.15}
\resizebox{\linewidth}{!}{
\begin{tabular}{r||c|c|c|c|c|c|c}
\spthickhline
\rowcolor[rgb]{0.92,0.92,0.92}
\textbf{Method}~~~~~~ & 1--2 & 2--3 & 3--4 & 4--5 & 5--6 & 6--7 & 7+ \\
\hline\hline
\rowcolor[rgb]{0.92,0.92,0.92}
\textbf{PKINet-v2-S}~~~ & \textbf{54.60} & \textbf{41.97} & \textbf{45.21} & \textbf{46.18} & \textbf{20.84} & \textbf{26.18} & \textbf{46.92} \\
PKINet-v1-S~\cite{cai2024poly} & 52.99 & 40.99 & 36.59 & 45.22 & 19.40 & 19.60 & 42.55 \\
Strip RCNN-S~\cite{yuan2025strip} & 53.33 & 40.30 & 41.16 & 45.01 & 18.52 & 20.44 & 44.59 \\
LSKNet-S~\cite{Li_2023_ICCV} & 54.41 & 41.11 & 36.19 & 41.64 & 19.40 & 24.43 & 43.11 \\
\spthickhline
\end{tabular}}
\vspace{-2pt}
\caption{Aspect-ratio bins}
\end{subtable}
\hfill
\begin{subtable}{0.496\linewidth}
\centering
\setlength{\tabcolsep}{5pt}
\renewcommand\arraystretch{1.15}
\resizebox{\linewidth}{!}{
\begin{tabular}{r||c|c|c|c|c}
\spthickhline
\rowcolor[rgb]{0.92,0.92,0.92}
\textbf{Method}~~~~~~ & 0.01--0.01 & 0.02--0.05 & 0.05--0.1 & 0.1--0.2 & 0.2+ \\
\hline\hline
\rowcolor[rgb]{0.92,0.92,0.92}
\textbf{PKINet-v2-S}~~~  & \textbf{13.43} & \textbf{35.69} & \textbf{54.82} & \textbf{57.30} & \textbf{71.39} \\
PKINet-v1-S~\cite{cai2024poly} & 12.98 & 34.32 & 53.49 & 55.32 & 69.50 \\
Strip RCNN-S~\cite{yuan2025strip} & 11.30 & 34.79 & 53.79 & 56.75 & 70.87 \\
LSKNet-S~\cite{Li_2023_ICCV} & 12.65 & 33.18 & 53.42 & 55.57 & 70.93 \\
\spthickhline
\end{tabular}}
\vspace{-2pt}
\caption{Object-size bins (relative area)}
\end{subtable}

\vspace{-16pt}
\end{table*}

\vspace{-20pt}
\subsection{Qualitative Results}
\label{sec:qua_result}
\begin{figure*}[t!]
    \centering    \includegraphics[width=\linewidth]{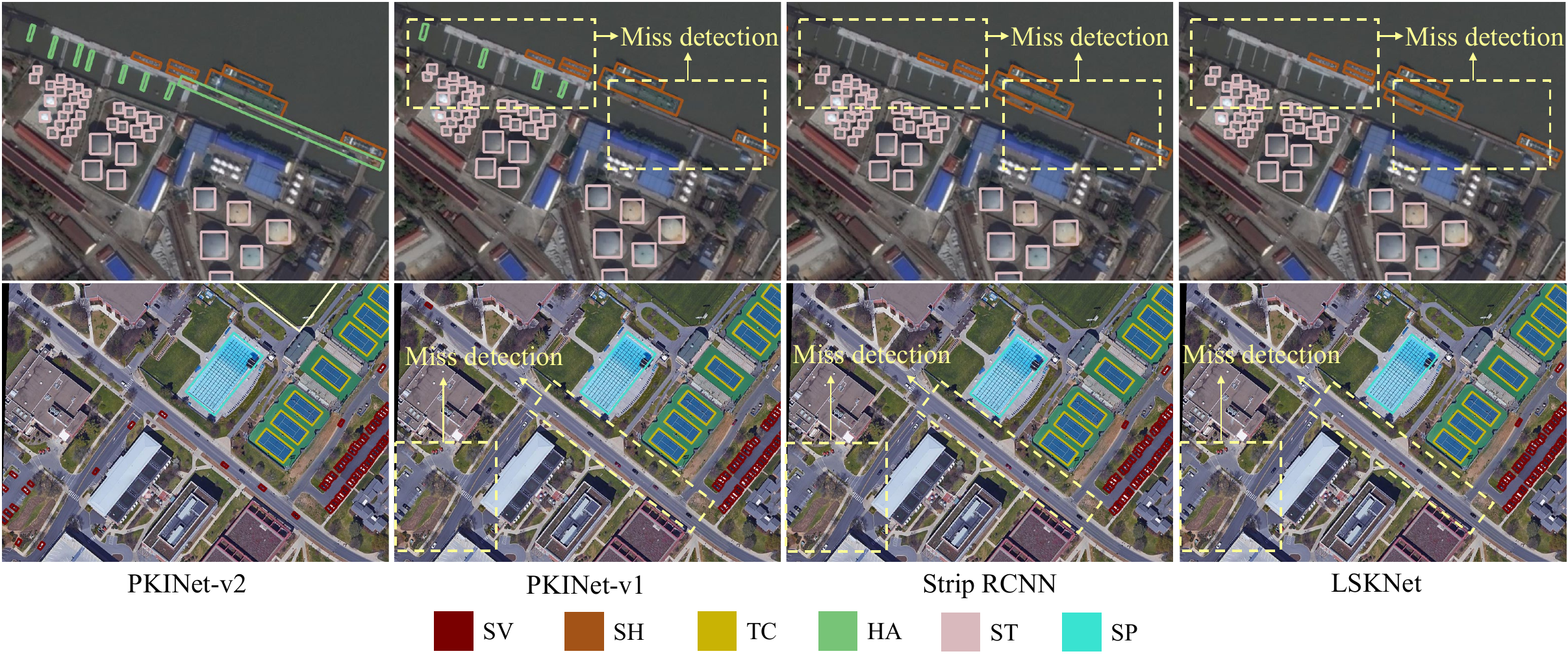}
    \vspace{-20pt}
    \caption{\textbf{Visual results on DOTA-1.0 dataset}~\cite{xia2018dota}. \textbf{(Top)} Geometric complexity. \textbf{(Bottom)} Spatial complexity. PKINet-v2 delivers more robust detection than other methods~\cite{cai2024poly,yuan2025strip,Li_2023_ICCV} under both challenges. Please refer to \S\ref{sec:qua_result} for more details.}
    \label{fig:vis}
    \vspace{-12pt}
\end{figure*}
Fig.~\ref{fig:vis} visualizes representative predictions on DOTA-v1.0~\cite{xia2018dota} under two challenges in RSIs. PKINet-v2 demonstrates a strong ability to adjust extreme aspect ratios (\textit{e.g.}, HA and SH) and large-scale variations (\textit{e.g.}, SP, TC, and SV).

\subsection{Diagnostic Experiments}
\label{sec:Dia_exp}
\begin{table*}[t]
\centering
\caption{\textbf{Diagnostic experiments on DOTA-v1.0}~\cite{xia2018dota}.
The adopted default designs are marked in {\color{red}red}.
All models are pretrained on ImageNet-1K~\cite{deng2009imagenet} for 100 epochs (ablation setting) and built within Oriented RCNN~\cite{xie2021oriented}. FPS reported in Tab.~\ref{tab:diag_kernel_design_v2} and Tab.~\ref{tab:diag_kernel_number_v2}  is measured during inference after applying HKR.
See \S\ref{sec:Dia_exp} for details.}
\vspace{-6pt}
\begin{subtable}{0.49\linewidth}
\resizebox{\textwidth}{!}{
\setlength\tabcolsep{2.5pt}
\renewcommand\arraystretch{1}
\begin{tabular}{c||c|c|c}
\spthickhline
\rowcolor[rgb]{0.92,0.92,0.92}
Kernel Design & \#\textbf{Params}$\downarrow$ & \textbf{FPS}$\uparrow$ & \textbf{mAP}$\uparrow$ \\
\hline\hline

Dense-only (3)                 & 13.2M & 54.6 & 78.57 \\ \hline
Strip-only (19)                & 13.3M & 54.6 & 79.62 \\ \hline
Sparse-only (3,5,7)            & 13.5M & 54.6 & 79.36 \\ \hline
{\color{red}Hybrid}            & 13.6M & 54.6 & 80.11 \\ \hline

\end{tabular}}
\vspace{-2pt}
\caption{kernel design}
\label{tab:diag_kernel_design_v2}
\end{subtable}
\hspace{-0.6em}
\begin{subtable}{0.49\linewidth}
\resizebox{\textwidth}{!}{
\setlength\tabcolsep{4.5pt}
\renewcommand\arraystretch{1.09}
\begin{tabular}{c||c|c|c}
\spthickhline
\rowcolor[rgb]{0.92,0.92,0.92}
Kernel Dilations & \#\textbf{Params}$\downarrow$ & \textbf{Max RF} & \textbf{mAP}$\uparrow$ \\
\hline\hline

(1,1,1,1)        & 13.6M & 7  & 79.88 \\ \hline
(1,2,2,2)        & 13.6M & 13 & 79.98 \\ \hline
{\color{red}(1,3,3,3)} & 13.6M & 19 & 80.11 \\ \hline
(1,4,4,4)        & 13.6M & 25 & 80.01 \\ \hline

\end{tabular}}
\vspace{-2pt}
\caption{dilations of square branches}
\label{tab:diag_kernel_dilation_v2}
\end{subtable}

\vspace{-10pt}

\begin{subtable}{0.49\linewidth}
\resizebox{\textwidth}{!}{
\setlength\tabcolsep{3pt}
\renewcommand\arraystretch{1.0}
\begin{tabular}{c||c|c|c}
\spthickhline
\rowcolor[rgb]{0.92,0.92,0.92}
Kernel Number & \#\textbf{Params}$\downarrow$ & \textbf{FPS}$\uparrow$ & \textbf{mAP}$\uparrow$ \\
\hline\hline

1(19)            & 13.3M & 54.6 & 79.62 \\ \hline
3(5,7,19)        & 13.5M & 54.6 & 79.90 \\ \hline
4(3,5,7,19)      & 13.5M & 54.6 & 79.99 \\ \hline
{\color{red}5}(3,3,5,7,19) & 13.6M & 54.6 & 80.11 \\ \hline

\end{tabular}}
\vspace{-2pt}
\caption{number of parallel DW branches in PKS}
\label{tab:diag_kernel_number_v2}
\end{subtable}
\hspace{-0.6em}
\begin{subtable}{0.49\linewidth}
\resizebox{\textwidth}{!}{
\setlength\tabcolsep{4.5pt}
\renewcommand\arraystretch{1.02}
\begin{tabular}{c|c||c|c|c}
\spthickhline
\rowcolor[rgb]{0.92,0.92,0.92}
Variant & HKR & \textbf{FPS}$\uparrow$ & $\Delta$\textbf{FPS} & \textbf{mAP}$\uparrow$\\ \hline\hline

T & w/o HKR            & 46.2 & --   & 79.10 \\ \hline
T & {\color{red}w/ HKR} & 58.0 & 11.8 & 79.10 \\ \hline
S & w/o HKR            & 43.4 & --   & 80.11 \\ \hline
S & {\color{red}w/ HKR} & 54.6 & 11.2 & 80.11 \\ \hline

\end{tabular}}
\vspace{-2pt}
\caption{effect of HKR re-parameterization}
\label{tab:diag_hkr_v2}
\end{subtable}
\vspace{-18pt}
\label{tab:diagnostic_v2}
\vspace{-8pt}
\end{table*}

We conduct diagnostic experiments on DOTA-v1.0 using Oriented RCNN~\cite{xie2021oriented}. All backbones are pretrained on ImageNet-1K~\cite{deng2009imagenet} for 100 epochs for efficiency.

\noindent\textbf{Kernel Design.}
Tab.~\ref{tab:diag_kernel_design_v2} shows that dense-only kernels perform worst due to limited long-range modeling. Strip-only and sparse-only yield sub-optimal mAP, while the hybrid design achieves the best mAP, indicating the necessity of jointly synergizing anisotropic strip convolutions with isotropic square convolutions.

\noindent\textbf{Kernel Dilations.}
Varying dilations of square branches (Tab.~\ref{tab:diag_kernel_dilation_v2}), the default (1,3,3,3) is optimal. Smaller or larger dilations slightly degrade accuracy.

\noindent\textbf{Kernel Number.}
Increasing the number of parallel DW branches (Tab.~\ref{tab:diag_kernel_number_v2}) consistently improves mAP, and the 5-branch setting reaches the best result.

\noindent\textbf{HKR Re-parameterization.}
Tab.~\ref{tab:diag_hkr_v2} shows that HKR keeps mAP unchanged but largely boosts FPS on both variants (T: 46.2$\rightarrow$58.0, +11.8; S: 43.4$\rightarrow$54.6, +11.2), validating its efficiency gain without accuracy loss.

\section{Conclusion}
In this paper, we propose Poly Kernel Inception Network v2 (PKINet-v2) for remote sensing object detection, which aims to simultaneously tackle the challenges of geometric and spatial complexity in remote sensing images. 
Distinct from prior methods, PKINet-v2 synergizes anisotropic axial-strip convolutions with isotropic square kernels and constructs a multi-scope receptive field, enabling robust modeling for both slender targets and regular-shaped objects across various scales. 
To reconcile model complexity with inference efficiency, a Heterogeneous Kernel Re-parameterization (HKR) Strategy is introduced by transforming the multi-branch architecture into a hardware-friendly single-branch structure. 
Extensive experiments demonstrate that PKINet-v2 achieves state-of-the-art performance on four widely-used benchmarks, while also achieving the fastest inference speed and a 3.9$\times$ inference speedup over PKINet-v1.

\clearpage
\bibliographystyle{splncs04}
\bibliography{main}

\clearpage
\setcounter{page}{1}
\setcounter{section}{0}
\setcounter{table}{0}
\setcounter{figure}{0}
\renewcommand{\thetable}{S\arabic{table}}
\renewcommand{\thefigure}{S\arabic{figure}}
\maketitlesupplementary

This appendix contains additional details for the ECCV 2026 submission, titled \textit{PKINet-v2: Towards Powerful and Efficient Poly-Kernel Remote Sensing Object Detection}. The appendix provides additional experimental results, more details of our approach and discussion, organized as follows:
\begin{itemize}[leftmargin=0.5cm]
        \item \S\ref{sec:social_impacts}: Social impacts
        \item \S\ref{sec:limitations}: Limitations and future work
        \item \S\ref{sec:more_quan_results}: More quantitative results
        \item \S\ref{sec:more_qua_results}: More qualitative results
        \item \S\ref{sec:pseudo_code}: Pseudo Code of Our Algorithm
\end{itemize}

\section{Social Impacts}
\label{sec:social_impacts}
This work presents PKINet-v2, a new backbone for remote sensing object detection that improves both effectiveness and efficiency by jointly addressing geometric and spatial complexity and by accelerating inference via the HKR re-parameterization strategy.
On the positive side, stronger and faster detection in remote sensing imagery may benefit many civilian applications, such as environmental and ecological monitoring, urban planning, disaster response and damage assessment, infrastructure inspection, maritime management, and precision agriculture.
In particular, the improved inference throughput can facilitate large-area and near-real-time analysis, and may reduce the compute and energy cost per processed image when deployed at scale.

On the negative side, remote sensing detection technology may be misused for surveillance or military purposes (e.g., large-scale monitoring or targeting).
Moreover, the approach still relies on the data distributions and annotations of available datasets; domain shifts across regions, sensors, seasons, or imaging conditions can introduce bias and inaccuracies, which may lead to harmful downstream decisions if outputs are treated as definitive.
To mitigate potential adverse societal impacts, we recommend responsible-use practices, including strict access control and security protocols for deployed systems, transparent reporting of failure cases and limitations, dataset-aware bias evaluation across diverse geographies, and verification for high-stakes applications.

\section{Limitations and Future Work}
\label{sec:limitations}
Despite PKINet-v2 delivering state-of-the-art detection accuracy while substantially improving inference efficiency, there remain several limitations:

\begin{itemize}[leftmargin=1.2em, itemsep=2pt, topsep=2pt]
    \item \textbf{Scaling up model capacity.}
    Due to constrained computational resources, we mainly investigate the Tiny/Small variants and do not fully explore scaling up the model capacity to unlock the maximal potential of the proposed hybrid multi-scope design.
    Systematic scalability studies (e.g., larger width/depth, stronger pretraining, and larger-scale training regimes) have shown clear gains in general vision backbones, as exemplified by Swin Transformer~\cite{liu2021swin} and ConvNeXt~\cite{liu2022convnet}.
    We leave a comprehensive investigation of PKINet-v2 scalability for future work.

    \item \textbf{Extending to broader remote sensing tasks.}
    Our experiments focus on remote sensing \emph{object detection} benchmarks with oriented annotations.
    Although PKINet-v2 is designed as a general-purpose feature extractor and HKR can potentially benefit deployment across tasks, we have not yet extended it to broader remote sensing applications such as scene classification, semantic segmentation, and change detection.
    Evaluating and adapting PKINet-v2 under task-specific heads, supervision signals, and resolution requirements is an important direction for future research.
\end{itemize}

\section{More Quantitative Results}
\label{sec:more_quan_results}
Tab.~\ref{tab:var_arch2} reports the detailed performance of PKINet-v2 when integrated into various detection frameworks on DOTA-v1.0~\cite{xia2018dota}. PKINet-v2 consistently improves detection performance across all frameworks, demonstrating strong generality and robustness.

\begin{table*}[h]
 \vspace{-5pt}
\setlength{\tabcolsep}{2pt}
\renewcommand\arraystretch{1.15}
\scriptsize
\centering
\caption{\textbf{Experimental results on DOTA-v1.0 dataset}~\cite{xia2018dota} across
different detection frameworks under the single-scale training and testing setting.
PKINet-v2 is pretrained on ImageNet-1K~\cite{deng2009imagenet} for 300 epochs similar to previous methods~\cite{cai2024poly,Li_2023_ICCV,yuan2025strip}.}
\resizebox{\textwidth}{!}{
\hspace{-1.0em}
\begin{tabular}{r|c|c||c|ccccccccccccccc}
\thickhline
\rowcolor[rgb]{0.92,0.92,0.92}
Method & \textbf{Backbone} & \textbf{mAP} $\uparrow$ & \#\textbf{P} $\downarrow$
& PL & BD & BR & GTF & SV & LV & SH & TC & BC & ST & SBF & RA & HA & SP & HC \\
\hline
\hline
\multicolumn{3}{l}{\textbf{\textit{DETR-based}}}\\
\hline
AO\textsuperscript{2}-DETR~\cite{dai2022ao2}
& ResNet-50~\cite{he2016deep} & 70.91 & 40.8M
& 87.99 & 79.46 & 45.74 & 66.64 & 78.90 & 73.90 & 73.30 & 90.40 & 80.55 & 85.89 & 55.19 & 63.62 & 51.83 & 70.15 & 60.04 \\
\hline
O\textsuperscript{2}-DETR~\cite{ma2021oriented}
& ResNet-50~\cite{he2016deep} & 72.15 & -
& 86.01 & 75.92 & 46.02 & 66.65 & 79.70 & 79.93 & 89.17 & 90.44 & 81.19 & 76.00 & 56.91 & 62.45 & 64.22 & 65.80 & 58.96 \\
\hline
ARS-DETR~\cite{zeng2023ars}
& ResNet-50~\cite{he2016deep} & 73.79 & 41.6M
& 86.61 & 77.26 & 48.84 & 66.76 & 78.38 & 78.96 & 87.40 & 90.61 & 82.76 & 82.19 & 54.02 & 62.61 & 72.64 & 72.80 & 64.96 \\
\hline
\hline
\multicolumn{3}{l}{\textbf{\textit{One-stage}}}\\
\hline
SASM~\cite{hou2022shape}
& ResNet-50~\cite{he2016deep} & 74.92 & 36.6M
& 86.42 & 78.97 & 52.47 & 69.84 & 77.30 & 75.99 & 86.72 & 90.89 & 82.63 & 85.66 & 60.13 & 68.25 & 73.98 & 72.22 & 62.37 \\
\hline
R3Det-GWD~\cite{yang2021rethinking}
& ResNet-50~\cite{he2016deep} & 76.34 & 41.9M
& 88.82 & 82.94 & 55.63 & 72.75 & 78.52 & 83.10 & 87.46 & 90.21 & 86.36 & 85.44 & 64.70 & 61.41 & 73.46 & 76.94 & 57.38 \\
\hline
R3Det-KLD~\cite{yang2021learning}
& ResNet-50~\cite{he2016deep} & 77.36 & 41.9M
& 88.90 & 84.17 & 55.80 & 69.35 & 78.72 & 84.08 & 87.00 & 89.75 & 84.32 & 85.73 & 64.74 & 61.80 & 76.62 & 78.49 & 70.89 \\
\hline
O-RepPoints~\cite{li2022oriented}
& ResNet-50~\cite{he2016deep} & 75.97 & 36.6M
& 87.02 & 83.17 & 54.13 & 71.16 & 80.18 & 78.40 & 87.28 & 90.90 & 85.97 & 86.25 & 59.90 & 70.49 & 73.53 & 72.27 & 58.97 \\
\hline
\multirow{3}{*}{\parbox{2cm}{\raggedleft Rotated\\ FCOS~\cite{tian2019fcos}}}
& ResNet-50~\cite{he2016deep} & 72.45 & 31.9M
& 88.52 & 77.54 & 47.06 & 63.78 & 80.42 & 80.50 & 87.34 & 90.39 & 77.83 & 84.13 & 55.45 & 65.84 & 66.02 & 72.77 & 49.17 \\
& PKINet-v1-S~\cite{cai2024poly} & 74.86 & 21.7M
& 88.56 & 82.89 & 47.96 & 58.20 & 81.09 & 83.09 & 88.23 & 90.88 & 84.57 & 85.81 & 57.98 & 66.26 & 75.12 & 80.93 & 51.39 \\
& \textbf{PKINet-v2-S} & \textbf{77.13} & 21.6M
& 89.21 & 84.07 & 52.69 & 64.61 & 82.07 & 84.60 & 88.39 & 90.85 & 86.97 & 85.81 & 62.77 & 65.17 & 76.52 & 82.09 & 61.14 \\
\hdashline
\multirow{4}{*}{R3Det~\cite{yang2021r3det}}
& ResNet-50~\cite{he2016deep} & 69.70 & 41.9M
& 89.00 & 75.60 & 46.64 & 67.09 & 76.18 & 73.40 & 79.02 & 90.88 & 78.62 & 84.88 & 59.00 & 61.16 & 63.65 & 62.39 & 37.94 \\
& ARC-R50~\cite{pu2023adaptive} & 72.32 & 65.2M
& 89.49 & 78.04 & 46.36 & 68.89 & 77.45 & 72.87 & 82.76 & 90.90 & 83.07 & 84.89 & 58.72 & 68.61 & 64.75 & 68.39 & 49.67 \\
& PKINet-v1-S~\cite{cai2024poly} & 75.89 & 28.1M
& 89.63 & 82.40 & 49.77 & 71.72 & 79.95 & 81.39 & 87.79 & 90.90 & 84.20 & 86.09 & 61.08 & 66.55 & 73.06 & 73.85 & 59.95 \\
& \textbf{PKINet-v2-S} & \textbf{77.64} & 28.0M
& 90.06 & 84.81 & 54.96 & 74.72 & 79.89 & 82.81 & 87.65 & 90.75 & 87.26 & 86.06 & 66.40 & 65.24 & 75.24 & 80.74 & 58.04 \\
\hdashline
\multirow{4}{*}{S$^2$ANet~\cite{han2021align}}
& ResNet-50~\cite{he2016deep} & 74.12 & 38.5M
& 89.11 & 82.84 & 48.37 & 71.11 & 78.11 & 78.39 & 87.25 & 90.83 & 84.90 & 85.64 & 60.36 & 62.60 & 65.26 & 69.13 & 57.94 \\
& ARC-R50~\cite{pu2023adaptive} & 75.49 & 71.8M
& 89.28 & 78.77 & 53.00 & 72.44 & 79.81 & 77.84 & 86.81 & 90.88 & 84.27 & 86.20 & 60.74 & 68.97 & 66.35 & 71.25 & 65.77 \\
& PKINet-v1-S~\cite{cai2024poly} & 77.83 & 24.8M
& 89.67 & 84.16 & 51.94 & 71.89 & 80.81 & 83.47 & 88.29 & 90.80 & 87.01 & 86.94 & 65.02 & 69.53 & 75.83 & 80.20 & 61.85 \\
& \textbf{PKINet-v2-S} & \textbf{79.68} & 24.7M
& 89.82 & 86.35 & 56.65 & 76.58 & 81.58 & 85.07 & 88.27 & 90.88 & 88.01 & 86.40 & 68.12 & 67.05 & 77.62 & 79.74 & 73.07 \\
\hline
\hline
\multicolumn{3}{l}{\textbf{\textit{Two-stage}}}\\
\hline
SCRDet~\cite{yang2019scrdet}
& ResNet-50~\cite{he2016deep} & 72.61 & 41.9M
& 89.98 & 80.65 & 52.09 & 68.36 & 68.36 & 60.32 & 72.41 & 90.85 & 87.94 & 86.86 & 65.02 & 66.68 & 66.25 & 68.24 & 65.21 \\
\hline
G.V.~\cite{xu2020gliding}
& ResNet-50~\cite{he2016deep} & 75.02 & 41.1M
& 89.64 & 85.00 & 52.26 & 77.34 & 73.01 & 73.14 & 86.82 & 90.74 & 79.02 & 86.81 & 59.55 & 70.91 & 72.94 & 70.86 & 57.32 \\
\hline
CenterMap~\cite{long2021creating}
& ResNet-50~\cite{he2016deep} & 71.59 & 41.1M
& 89.02 & 80.56 & 49.41 & 61.98 & 77.99 & 74.19 & 83.74 & 89.44 & 78.01 & 83.52 & 47.64 & 65.93 & 63.68 & 67.07 & 61.59 \\
\hline
ReDet~\cite{han2021redet}
& ResNet-50~\cite{he2016deep} & 76.25 & 31.6M
& 88.79 & 82.64 & 53.97 & 74.00 & 78.13 & 84.06 & 88.04 & 90.89 & 87.78 & 85.75 & 61.76 & 60.39 & 75.96 & 68.07 & 63.59 \\
\hline
\multirow{3}{*}{RoI Trans.~\cite{ding2019learning}}
& ResNet-50~\cite{he2016deep} & 74.05 & 55.1M
& 89.01 & 77.48 & 51.64 & 72.07 & 74.43 & 77.55 & 87.76 & 90.81 & 79.71 & 85.27 & 58.36 & 64.11 & 76.50 & 71.99 & 54.06 \\
& PKINet-v1-S~\cite{cai2024poly} & 77.17 & 44.8M
& 89.33 & 85.59 & 55.75 & 74.69 & 74.69 & 79.13 & 88.05 & 90.90 & 87.43 & 86.90 & 61.67 & 64.25 & 77.77 & 75.38 & 66.08 \\
& \textbf{PKINet-v2-S} & \textbf{79.87} & 44.7M
& 89.47 & 85.98 & 58.90 & 79.42 & 74.53 & 85.68 & 88.04 & 90.87 & 87.78 & 87.14 & 68.50 & 63.26 & 78.99 & 81.78 & 77.76 \\
\hdashline
\multirow{4}{*}{\parbox{2cm}{\raggedleft Rotated Faster\\ R-CNN~\cite{ren2015faster}}}
& ResNet-50~\cite{he2016deep} & 73.17 & 41.1M
& 89.40 & 81.81 & 47.28 & 67.44 & 73.96 & 73.12 & 85.03 & 90.90 & 85.15 & 84.90 & 56.60 & 64.77 & 64.70 & 70.28 & 62.22 \\
& ARC-R50~\cite{pu2023adaptive} & 74.77 & 74.4M
& 89.49 & 82.11 & 51.02 & 70.38 & 79.07 & 75.06 & 86.18 & 90.91 & 84.23 & 86.41 & 56.10 & 69.42 & 65.87 & 71.90 & 63.47 \\
& PKINet-v1-S~\cite{cai2024poly} & 76.45 & 30.8M
& 89.33 & 85.27 & 52.34 & 73.03 & 73.72 & 75.60 & 86.97 & 90.88 & 86.52 & 87.30 & 64.23 & 64.20 & 75.63 & 80.31 & 61.47 \\
& \textbf{PKINet-v2-S} & \textbf{79.02} & 30.7M
& 89.67 & 86.39 & 58.00 & 78.49 & 80.09 & 77.23 & 87.26 & 90.83 & 88.37 & 87.36 & 65.84 & 64.42 & 77.71 & 80.25 & 73.38 \\
\hdashline
\multirow{6}{*}{O-RCNN~\cite{xie2021oriented}}
& ResNet-50~\cite{he2016deep} & 75.87 & 41.1M
& 89.46 & 82.12 & 54.78 & 70.86 & 78.93 & 83.00 & 88.20 & 90.90 & 87.50 & 84.68 & 63.97 & 67.69 & 74.94 & 68.84 & 52.28 \\
& ARC-R50~\cite{pu2023adaptive} & 77.35 & 74.4M
& 89.40 & 82.48 & 55.33 & 73.88 & 79.37 & 84.05 & 88.06 & 90.90 & 86.44 & 84.83 & 63.63 & 70.32 & 74.29 & 71.91 & 65.43 \\
& LSKNet-S~\cite{Li_2023_ICCV} & 77.49 & 31.0M
& 89.66 & 85.52 & 57.72 & 75.70 & 74.95 & 78.69 & 88.24 & 90.88 & 86.79 & 86.38 & 66.92 & 63.77 & 77.77 & 74.47 & 64.82 \\
& LWGANet-L2~\cite{lu2025lwganet} & 79.02 & 29.2M
& 89.49 & 85.48 & 54.93 & 77.12 & 81.59 & 85.64 & 88.43 & 90.85 & 87.23 & 86.78 & 67.47 & 65.06 & 78.23 & 73.33 & 73.66 \\
& PKINet-v1-S~\cite{cai2024poly} & 78.39 & 30.8M
& 89.72 & 84.20 & 55.81 & 77.63 & 80.25 & 84.45 & 88.12 & 90.88 & 87.57 & 86.07 & 66.86 & 70.23 & 77.47 & 73.62 & 62.94 \\
& \textbf{PKINet-v2-S} & \textbf{80.46} & 30.7M
& 89.53 & 85.19 & 57.72 & 78.63 & 81.21 & 86.38 & 88.18 & 90.84 & 87.95 & 86.25 & 71.97 & 67.76 & 78.15 & 81.80 & 75.34 \\
\hline
\end{tabular}}
\label{tab:var_arch2}
\end{table*}

\section{More Qualitative Results}
\label{sec:more_qua_results}

\begin{figure*}[h]
	\begin{center}
		\includegraphics[width = \linewidth]{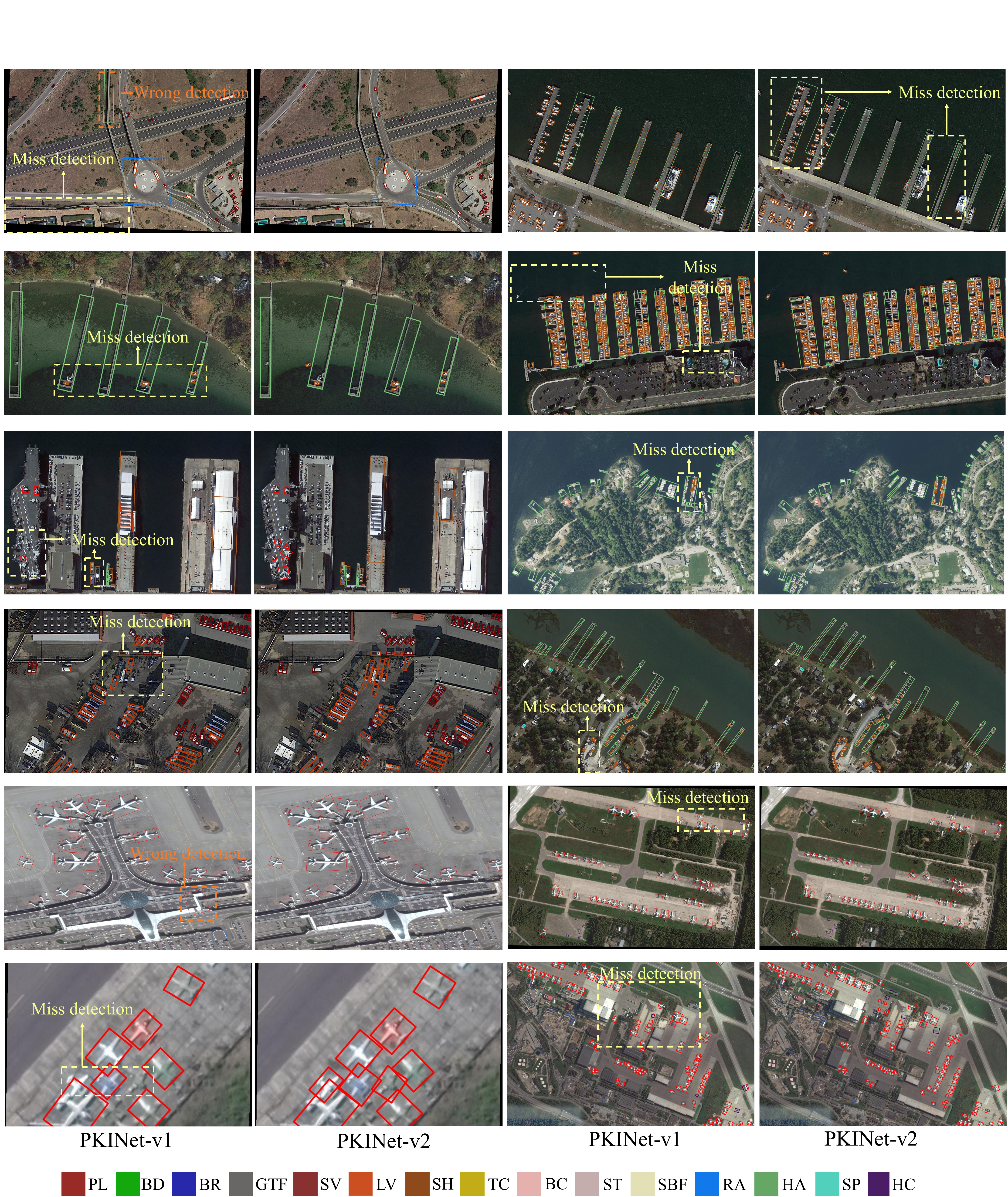}
	\end{center}
	\setlength{\abovecaptionskip}{0.cm}
	\caption{\textbf{More qualitative comparisons on DOTA} \cite{xia2018dota} on Oriented R-CNN \cite{xie2021oriented} with PKINet-v1~\cite{cai2024poly}. See \S\ref{sec:more_qua_results} for details.}
	\label{fig:orc_comparison}
\end{figure*}

\begin{figure*}[h]
	\begin{center}
		\includegraphics[width = \linewidth]{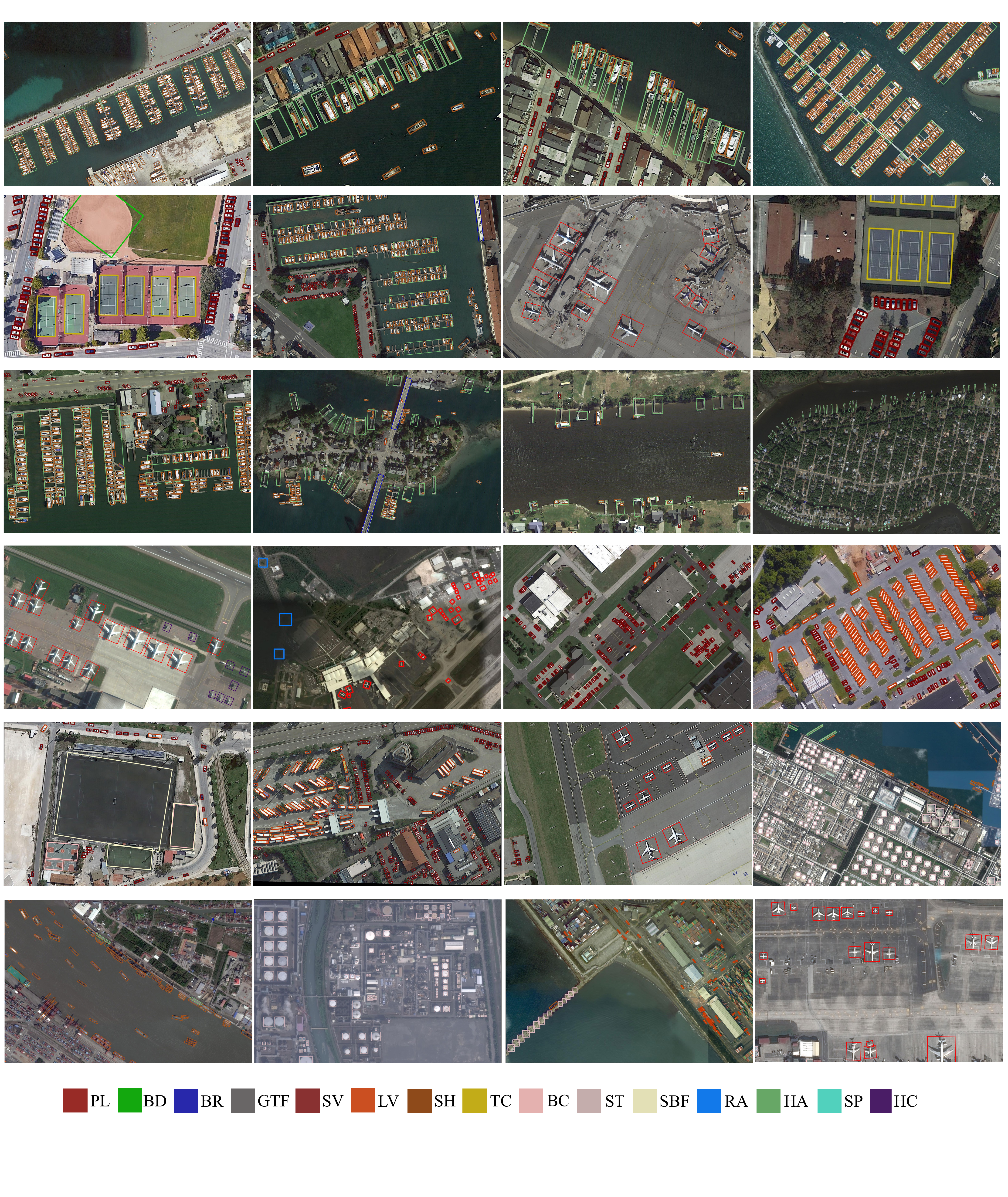}
	\end{center}
	\setlength{\abovecaptionskip}{0.cm}
        \vspace{-24pt}
	\caption{\textbf{More qualitative results on DOTA} \cite{xia2018dota} on Oriented R-CNN \cite{xie2021oriented} of PKINet-v2. See \S\ref{sec:more_qua_results} for details.}
	\label{fig:orc_successful}
\end{figure*}

In Fig.~\ref{fig:orc_comparison} and Fig.~\ref{fig:orc_successful}, we provide additional qualitative evidence on the DOTA benchmark~\cite{xia2018dota} under the Oriented R-CNN framework~\cite{xie2021oriented}. Fig.~\ref{fig:orc_comparison} presents side-by-side comparisons between PKINet-v1~\cite{cai2024poly} and PKINet-v2. Compared with PKINet-v1, PKINet-v2 yields more complete and accurate predictions for objects with diverse aspect ratios and arbitrary orientations across different scales, demonstrating superior robustness to geometric complexity. Fig.~\ref{fig:orc_successful} further illustrates that PKINet-v2 can reliably handle spatial complexity, including substantial scale variations from tiny instances to large objects and wide-ranging scene contexts. This advantage is particularly evident in challenging harbor and airport scenes with mixed object sizes, dense distributions, and strong background interference.

\begin{figure*}[t]
	\begin{center}
		\includegraphics[width = \linewidth]{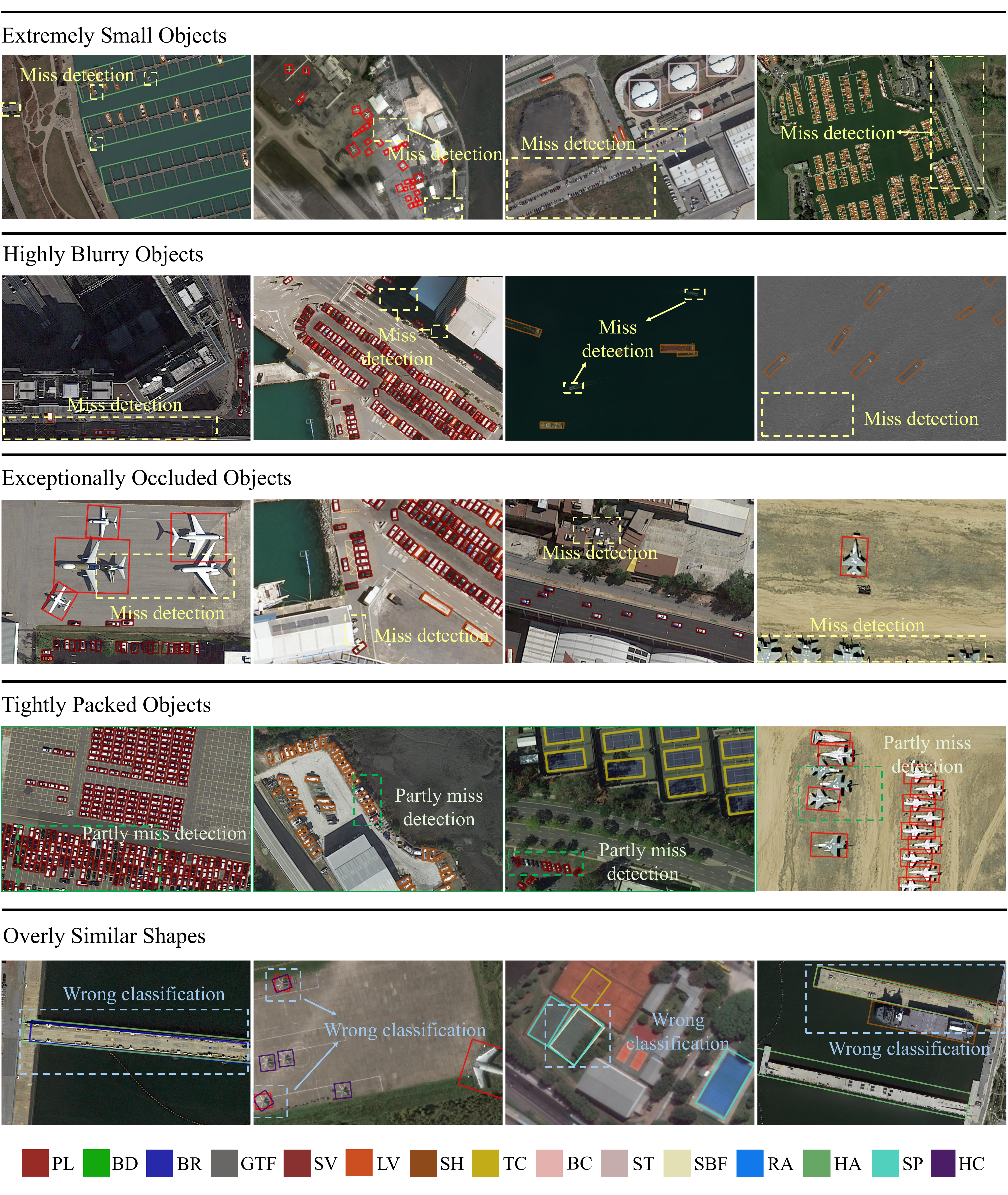}
	\end{center}
	\setlength{\abovecaptionskip}{0.cm}
	\caption{\textbf{Representative failure cases on DOTA} \cite{xia2018dota} over Oriented R-CNN \cite{xie2021oriented}. See \S\ref{sec:more_qua_results} for details.}
	\label{fig:failure_cases}
\end{figure*}

\noindent\textbf{Failure Case Analysis.} Despite the clear gains brought by our method, it still fails in a few highly challenging remote-sensing scenarios. Fig.~\ref{fig:failure_cases} summarizes representative failure patterns, which mainly fall into five categories: (i) extremely small objects, (ii) highly blurry objects, (iii) exceptionally occluded objects, (iv) tightly packed objects, and (v) overly similar shapes. We attribute these errors to the intrinsic difficulty of extracting reliable and discriminative features under such adverse conditions. In particular, extremely small targets contain limited pixels and weak cues, making them prone to being missed. Severe blur further suppresses edges and textures, leading to low-confidence responses and missed detections. Heavy occlusion disrupts object integrity and yields incomplete evidence, which can cause both localization failures and omissions. In crowded scenes, strong mutual overlap and ambiguous boundaries often result in partial misses. Finally, objects with highly similar appearances can confuse the classifier when shape/structure cues dominate, producing category-level confusion and wrong predictions.

\clearpage
\section{Pseudo Code of Our Algorithm}
\label{sec:pseudo_code}
This section provides a PyTorch-like pseudo-code to clarify the core components of PKINet-v2.
Alg.~\ref{alg:pks_train} describes the \emph{training-time} forward of a PKINet-v2 block, including the PKS Block and the underlying PKS Module with heterogeneous depth-wise branches (axial-strip and square kernels).
Alg.~\ref{alg:pks_infer} presents the \emph{inference-time} PKS Module after applying HKR, where all heterogeneous branches are collapsed into a single fused $K_{\max}\!\times\!K_{\max}$ depth-wise convolution for efficient deployment.
Alg.~\ref{alg:hkr} details the HKR procedure itself, including Conv--BN fusion and the heterogeneous-to-homogeneous kernel conversion that algebraically scatters each branch into the unified fused kernel while preserving identical outputs.

\begin{algorithm}[h]
\caption{Training-time forward of a PKINet-v2 Block.}
\label{alg:pks_train}
\definecolor{codeblue}{rgb}{0.25,0.5,0.5}
\lstset{
 backgroundcolor=\color{white},
 basicstyle=\fontsize{7.4pt}{7.4pt}\ttfamily\selectfont,
 columns=fullflexible,
 breaklines=true,
 escapeinside={(:}{:)},
 commentstyle=\fontsize{7.4pt}{7.4pt}\color{codeblue},
 keywordstyle=\fontsize{7.4pt}{7.4pt},
 frame=none,
}
\begin{lstlisting}[language=python]

def PKINetV2_Block_Forward(x, Norm1, PKS_Block, Norm2, FFN, DropPath, LS1, LS2):
    x = x + DropPath( LS1.view(1,-1,1,1) * PKS_Block( Norm1(x) ) )
    x = x + DropPath( LS2.view(1,-1,1,1) * FFN( Norm2(x) ) )
    return x


def PKS_Block_Forward(x, Proj1, GELU, PKS_Module, Proj2):
    shortcut = x
    x = Proj1(x)
    x = GELU(x)
    x = PKS_Module(x)          # spatial gating unit (RepStripModuleV6)
    x = Proj2(x)
    return x + shortcut        # residual inside PKS Block


def PKS_Module_Train(x,
                     Conv0_DW5,                    # self.conv0: DWConv 5x5
                     Axial_1x19, Axial_19x1, BN_ax, # branch1_axial: (1x19)->(19x1)->BN
                     DW7_d3, BN7,                  # branch2_sparse: 7x7, d=3 + BN
                     DW5_d3, BN5,                  # branch3_sparse: 5x5, d=3 + BN
                     DW3_d3, BN3s,                 # branch4_sparse: 3x3, d=3 + BN
                     DW3_d1, BN3d,                 # branch5_dense : 3x3, d=1 + BN
                     Conv1_1x1):                   # self.conv1: 1x1 mixing

    x_feat = Conv0_DW5(x)

    z_ax = BN_ax( Axial_19x1( Axial_1x19(x_feat) ) )
    z_7  = BN7 ( DW7_d3(x_feat) )
    z_5  = BN5 ( DW5_d3(x_feat) )
    z_3s = BN3s( DW3_d3(x_feat) )
    z_3d = BN3d( DW3_d1(x_feat) )

    attn = Conv1_1x1( z_ax + z_7 + z_5 + z_3s + z_3d )
    return x * attn
\end{lstlisting}
\end{algorithm}

\begin{algorithm}[h]
\caption{Inference-time PKS Module after HKR.}
\label{alg:pks_infer}
\definecolor{codeblue}{rgb}{0.25,0.5,0.5}
\lstset{
 backgroundcolor=\color{white},
 basicstyle=\fontsize{7.4pt}{7.4pt}\ttfamily\selectfont,
 columns=fullflexible,
 breaklines=true,
 escapeinside={(:}{:)},
 commentstyle=\fontsize{7.4pt}{7.4pt}\color{codeblue},
 keywordstyle=\fontsize{7.4pt}{7.4pt},
 frame=none,
}
\begin{lstlisting}[language=python]
def PKS_Module_Infer(x, DWConv5, DWConv_Kmax, Conv1x1, W_star, b_star):
    x_feat = DWConv5(x)
    attn   = DWConv_Kmax(x_feat, weight=W_star, bias=b_star)  # single fused DWConv
    attn   = Conv1x1(attn)
    return x * attn
\end{lstlisting}
\end{algorithm}

\begin{algorithm}[h]
\caption{HKR: Heterogeneous Kernel Re-parameterization.}
\label{alg:hkr}
\definecolor{codeblue}{rgb}{0.25,0.5,0.5}
\lstset{
 backgroundcolor=\color{white},
 basicstyle=\fontsize{7.4pt}{7.4pt}\ttfamily\selectfont,
 columns=fullflexible,
 breaklines=true,
 escapeinside={(:}{:)},
 commentstyle=\fontsize{7.4pt}{7.4pt}\color{codeblue},
 keywordstyle=\fontsize{7.4pt}{7.4pt},
 frame=none,
}
\begin{lstlisting}[language=python]
def fuse_conv_bn(conv_w, bn_running_mean, bn_running_var, bn_gamma, bn_beta, bn_eps):
    # conv_w: (C,1,k,k) or (C,1,1,k) or (C,1,k,1), bias is assumed 0 in training branches
    std = (bn_running_var + bn_eps).sqrt()                 # (C,)
    scale = (bn_gamma / std).view(-1, 1, 1, 1)             # (C,1,1,1)
    w_hat = conv_w * scale
    b_hat = bn_beta - bn_running_mean * bn_gamma / std     # (C,)
    return w_hat, b_hat


def scatter_square_into(W_star, w_small, k, d, Kmax):
    # w_small: (C,1,k,k), scatter by dilation-aware offsets into W_star: (C,1,Kmax,Kmax)
    c0 = Kmax // 2
    c1 = k // 2
    for i in range(k):
        for j in range(k):
            hi = c0 + (i - c1) * d
            wj = c0 + (j - c1) * d
            W_star[:, :, hi, wj] += w_small[:, :, i, j]


def HKR_Reparameterize(
    Kmax, C,
    # axial-strip branch (1xK then Kx1 then BN on the second conv output)
    w_1xK, w_Kx1, bn_ax,
    # square branches: (w_conv, bn, k, dilation)
    square_branches
):
    W_star = zeros((C, 1, Kmax, Kmax))
    b_star = zeros((C,))

    # (1) axial-strip branch: fuse BN into Kx1 conv, then outer product to get KxK
    w_Kx1_hat, b_ax = fuse_conv_bn(w_Kx1,
                                   bn_ax.mean, bn_ax.var, bn_ax.gamma, bn_ax.beta, bn_ax.eps)
    # channel-wise outer product: (C,1,K,1) x (C,1,1,K) -> (C,1,K,K)
    w_ax_eff = matmul(w_Kx1_hat, w_1xK)   # per-channel matmul
    scatter_square_into(W_star, w_ax_eff, k=Kmax, d=1, Kmax=Kmax)
    b_star += b_ax

    # (2) square branches (dense/sparse): fuse Conv+BN then scatter by dilation
    for (w_k, bn_k, k, d) in square_branches:
        w_hat, b_hat = fuse_conv_bn(w_k,
                                    bn_k.mean, bn_k.var, bn_k.gamma, bn_k.beta, bn_k.eps)
        scatter_square_into(W_star, w_hat, k=k, d=d, Kmax=Kmax)
        b_star += b_hat

    return W_star, b_star
\end{lstlisting}
\end{algorithm}

\end{document}